\documentclass[acmtog]{acmart}
\acmSubmissionID{316}

\usepackage{graphicx}
\usepackage{multirow}
\usepackage{makecell}
\usepackage{soul}

\usepackage{amsmath,amsfonts,bm}

\def\eqref#1{equation~\ref{#1}}

\def\1{\bm{1}}

\def\vd{{\bm{d}}}

\def\vf{{\bm{f}}}

\def\vp{{\bm{p}}}
\def\vq{{\bm{q}}}

\def\vt{{\bm{t}}}

\def\vw{{\bm{w}}}

\def\vz{{\bm{z}}}

\def\mI{{\bm{I}}}

\DeclareMathAlphabet{\mathsfit}{\encodingdefault}{\sfdefault}{m}{sl}
\SetMathAlphabet{\mathsfit}{bold}{\encodingdefault}{\sfdefault}{bx}{n}

\DeclareMathOperator*{\argmin}{arg\,min}

\DeclareMathOperator{\round}{round}

\renewcommand{\shortauthors}{Pan, et al.}

\newcommand{\eg}{\textit{e.g.}}
\newcommand{\ie}{\textit{i.e.}}

\newcommand{\etal}{\textit{et~al.}}
\newcommand{\etc}{\textit{etc}}

\usepackage{graphicx}
\acmJournal{TOG}

\copyrightyear{2023}
\acmYear{2023}
\setcopyright{rightsretained}
\acmConference[SIGGRAPH '23 Conference Proceedings]{Special Interest Group on Computer Graphics and Interactive Techniques Conference Conference Proceedings}{August 6--10, 2023}{Los Angeles, CA, USA}
\acmBooktitle{Special Interest Group on Computer Graphics and Interactive Techniques Conference Conference Proceedings (SIGGRAPH '23 Conference Proceedings), August 6--10, 2023, Los Angeles, CA, USA}
\acmDOI{10.1145/3588432.3591500}
\acmISBN{979-8-4007-0159-7/23/08}

\begin{document}
\title{Drag Your GAN: Interactive Point-based Manipulation on the Generative Image Manifold}

\author{Xingang Pan}
\email{xpan@mpi-inf.mpg.de}
\affiliation{
  \institution{Max Planck Institute for Informatics}
  \country{Germany}
}
\affiliation{
  \institution{Saarbrücken Research Center for Visual Computing, Interaction and AI}
  \country{Germany}
}
\author{Ayush Tewari}
\email{ayusht@mit.edu}
\affiliation{
  \institution{MIT CSAIL}
  \country{USA}
}
\author{Thomas Leimkühler}
\email{thomas.leimkuehler@mpi-inf.mpg.de}
\affiliation{
  \institution{Max Planck Institute for Informatics}
  \country{Germany}
}
\author{Lingjie Liu}
\email{lingjie.liu@seas.upenn.edu}
\affiliation{
  \institution{Max Planck Institute for Informatics}
  \country{Germany}
}
\affiliation{
  \institution{University of Pennsylvania}
  \country{USA}
}
\author{Abhimitra Meka}
\email{abhim@google.com}
\affiliation{
  \institution{Google AR/VR}
  \country{USA}
}
\author{Christian Theobalt}
\email{theobalt@mpi-inf.mpg.de}
\affiliation{
  \institution{Max Planck Institute for Informatics}
  \country{Germany}
}
\affiliation{
  \institution{Saarbrücken Research Center for Visual Computing, Interaction and AI}
  \country{Germany}
}

\renewcommand{\shortauthors}{X. Pan, A. Tewari, T. Leimkühler, L. Liu, A. Meka, C. Theobalt}

\begin{abstract}
  Synthesizing visual content that meets users' needs often requires flexible and precise controllability of the pose, shape, expression, and layout of the generated objects. 
  Existing approaches gain controllability of generative adversarial networks (GANs) via manually annotated training data or a prior 3D model, 
  which often lack flexibility, precision, and generality.
  In this work, we study a powerful yet much less explored way of controlling GANs, that is, to "drag" any points of the image to precisely reach target points in a user-interactive manner, as shown in Fig.1.
  To achieve this, we propose \textit{DragGAN}, which consists of two main components: 1) a feature-based motion supervision that drives the handle point to move towards the target position, and 2) a new point tracking approach that leverages the discriminative generator features to keep localizing the position of the handle points.
  Through \textit{DragGAN}, anyone can deform an image with precise control over where pixels go, thus manipulating the pose, shape, expression, and layout of diverse categories such as animals, cars, humans, landscapes, \etc.
  As these manipulations are performed on the learned generative image manifold of a GAN, they tend to produce realistic outputs even for challenging scenarios such as hallucinating occluded content and deforming shapes that consistently follow the object's rigidity.
  Both qualitative and quantitative comparisons demonstrate the advantage of \textit{DragGAN} over prior approaches in the tasks of image manipulation and point tracking.
  We also showcase the manipulation of real images through GAN inversion.
\end{abstract}



\begin{teaserfigure}
  \includegraphics[width=\textwidth]{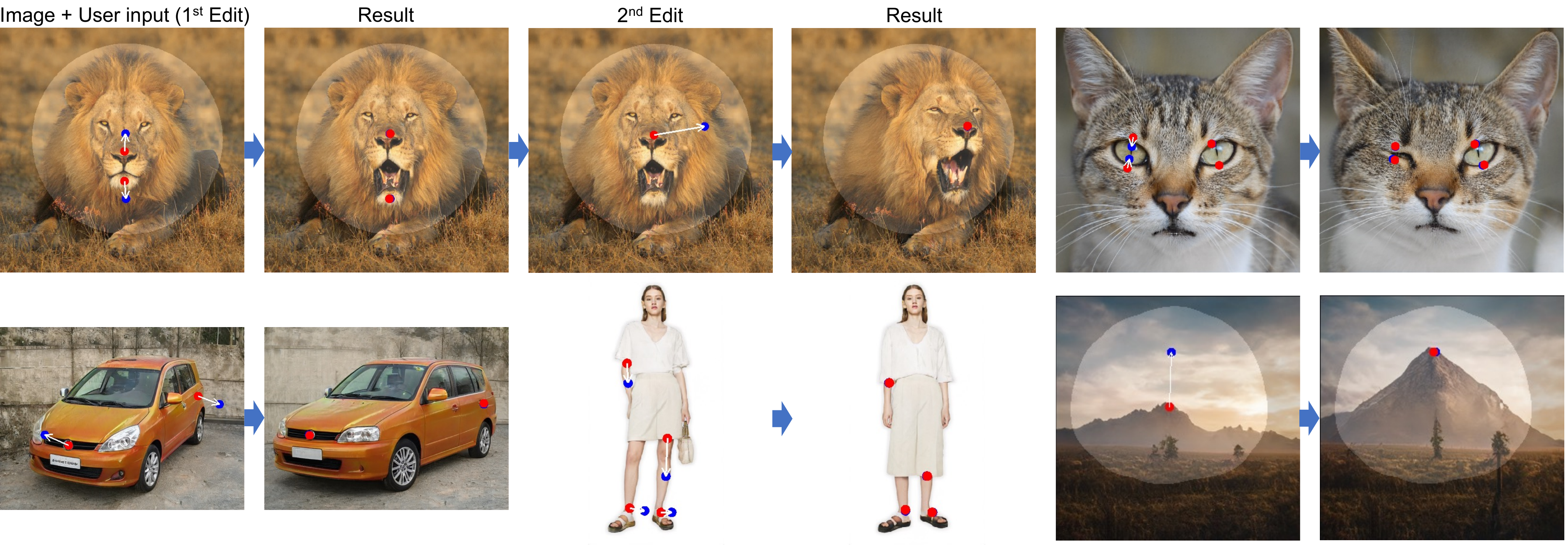}
  \vspace{-0.7cm}
  \caption{Our approach \textit{DragGAN} allows users to "drag" the content of any GAN-generated images. 
  Users only need to click a few handle points (\textcolor{red}{red}) and target points (\textcolor{blue}{blue}) on the image, and our approach will move the handle points to precisely reach their corresponding target points. 
  Users can optionally draw a mask of the flexible region (\textcolor{gray}{brighter} area), keeping the rest of the image fixed.
  This flexible point-based manipulation enables control of many spatial attributes like pose, shape, expression, and layout across diverse object categories.
  Project page: \textcolor{magenta}{\href{https://vcai.mpi-inf.mpg.de/projects/DragGAN/}{https://vcai.mpi-inf.mpg.de/projects/DragGAN/}}.
  }
  \label{fig:teaser}
\end{teaserfigure}

\maketitle

\section{Introduction}

Deep generative models such as generative adversarial networks (GANs)~\cite{goodfellow2014generative} have achieved unprecedented success in synthesizing random photorealistic images.
In real-world applications, a critical functionality requirement of such learning-based image synthesis methods is the controllability over the synthesized visual content.
For example, social-media users might want to adjust the position, shape, expression, and body pose of a human or animal in a casually-captured photo;
professional movie pre-visualization and media editing may require efficiently creating sketches of scenes with certain layouts;
and car designers may want to interactively modify the shape of their creations.
To satisfy these diverse user requirements, an \textit{ideal} controllable image synthesis approach should possess the following properties
1) \textit{Flexibility}: it should be able to control different spatial attributes including position, pose, shape, expression, and layout of the generated objects or animals; 
2) \textit{Precision}: it should be able to control the spatial attributes with high precision;
3) \textit{Generality}: it should be applicable to different object categories but not limited to a certain category.
While previous works only satisfy one or two of these properties, we target to achieve them all in this work.

Most previous approaches gain controllability of GANs via prior 3D models~\cite{tewari2020stylerig,deng2020disentangled,ghosh2020gif} or supervised learning that relies on manually annotated data~\cite{isola2017image,park2019semantic,ling2021editgan,shen2020interpreting,abdal2021styleflow}.
Thus, these approaches fail to generalize to new object categories, often control a limited range of spatial attributes or provide little control over the editing process.
Recently, text-guided image synthesis has attracted attention~\cite{rombach2021highresolution,saharia2022photorealistic,ramesh2022hierarchical}.
However, text guidance lacks precision and flexibility in terms of editing spatial attributes.
For example, it cannot be used to move an object by a specific number of pixels.

To achieve flexible, precise, and generic controllability of GANs, in this work, we explore a powerful yet much less explored interactive point-based manipulation.
Specifically, we allow users to click any number of handle points and target points on the image and the goal is to drive the handle points to reach their corresponding target points.
As shown in Fig.~\ref{fig:teaser}, this point-based manipulation allows users to control diverse spatial attributes 
and is agnostic to object categories.
The approach with the closest setting to ours is UserControllableLT~\cite{endoPG2022}, which also studies dragging-based manipulation.
Compared to it, the problem studied in this paper has two more challenges: 1) we consider the control of more than one point, which their approach does not handle well; 2) we require the handle points to precisely reach the target points while their approach does not.
As we will show in experiments, handling more than one point with precise position control enables much more diverse and accurate image manipulation.

To achieve such interactive point-based manipulation, we propose \textit{DragGAN}, which addresses two sub-problems, including 1) supervising the handle points to move towards the targets and 2) tracking the handle points so that their positions are known at each editing step.
Our technique is built on the key insight that the feature space of a GAN is sufficiently discriminative to enable both motion supervision and precise point tracking.
Specifically, the motion supervision is achieved via a shifted feature patch loss that optimizes the latent code.
Each optimization step leads to the handle points shifting closer to the targets; thus point tracking is then performed through nearest neighbor search in the feature space.
This optimization process is repeated until the handle points reach the targets.
\textit{DragGAN} also allows users to optionally draw a region of interest to perform region-specific editing.
Since \textit{DragGAN} does not rely on any additional networks like RAFT~\cite{teed2020raft}, it achieves efficient manipulation, only taking a few seconds on a single RTX 3090 GPU in most cases.
This allows for live, interactive editing sessions, in which the user can quickly iterate on different layouts till the desired output is achieved.

We conduct an extensive evaluation of \textit{DragGAN} on diverse datasets including animals (lions, dogs, cats, and horses), humans (face and whole body), cars, and landscapes. 
As shown in Fig.1, our approach effectively moves the user-defined handle points to the target points, achieving diverse manipulation effects across many object categories.
Unlike conventional shape deformation approaches that simply apply warping~\cite{igarashi2005rigid}, our deformation is performed on the learned image manifold of a GAN, which tends to obey the underlying object structures.
For example, our approach can hallucinate occluded content, like the teeth inside a lion's mouth, and can deform following the object's rigidity, like the bending of a horse leg.
We also develop a GUI for users to interactively perform the manipulation by simply clicking on the image.
Both qualitative and quantitative comparison confirms the advantage of our approach over UserControllableLT.
Furthermore, our GAN-based point tracking algorithm also outperforms existing point tracking approaches such as RAFT~\cite{teed2020raft} and PIPs~\cite{harley2022particle} for GAN-generated frames.
Furthermore, by combining with GAN inversion techniques, our approach also serves as a powerful tool for real image editing.
\section{Related Work}

\subsection{Generative Models for Interactive Content Creation}

Most current methods use generative adversarial networks (GANs) or diffusion models for controllable image synthesis. 

\paragraph{Unconditional GANs} GANs are generative models that 
transform low-dimensional randomly sampled latent vectors into photorealistic images. 
They are trained using adversarial learning and can be used to generate high-resolution photorealistic images~\cite{karras2019style,Karras2021,goodfellow2014generative,creswell2018generative}. 
Most GAN models like StyleGAN~\cite{karras2019style} do not directly enable controllable editing of the generated images.
\vspace{-2pt}
\paragraph{Conditional GANs} Several methods have proposed conditional GANs to address this limitation. 
Here, the network receives  a conditional input, such as segmentation map~\cite{isola2017image,park2019semantic} or 3D variables~\cite{deng2020disentangled,ghosh2020gif}, in addition to the randomly sampled latent vector to generate photorealistic images.
Instead of modeling the conditional distribution, EditGAN~\cite{ling2021editgan} enables editing by first modeling a joint distribution of images and segmentation maps, and then computing new images corresponding to edited segmentation maps.   
\vspace{-2pt}
\paragraph{Controllability using Unconditional GANs} 
Several methods have been proposed for editing unconditional GANs by manipulating the input latent vectors. 
Some approaches find meaningful latent directions via supervised learning from manual annotations or prior 3D models~\cite{tewari2020stylerig,shen2020interpreting,abdal2021styleflow,patashnik2021styleclip,FreeStyleGAN2021}.
Other approaches compute the important semantic directions in the latent space in an unsupervised manner~\cite{shen2020closed,harkonen2020ganspace,zhu2023linkgan}.
Recently, the controllability of coarse object position is achieved by introducing intermediate ``blobs"~\cite{epstein2022blobgan} or heatmaps~\cite{wang2022improving}.
All of these approaches enable editing of either image-aligned semantic attributes such as appearance, or coarse geometric attributes such as object position and pose.
While Editing-in-Style~\cite{collins2020editing} showcases some spatial attributes editing capability, it can only achieve this by transferring local semantics between different samples.
In contrast to these methods, our approach allows users to perform fine-grained control over the spatial attributes using point-based editing.

GANWarping~\cite{wang2022rewriting} also use point-based editing, however, they only enable out-of-distribution image editing. A few warped images can be used to update the generative model such that all generated images demonstrate similar warps. 
However, this method does not ensure that the warps lead to realistic images. Further, it does not enable controls such as changing the 3D pose of the object. 
Similar to us, UserControllableLT~\cite{endoPG2022} enables point-based editing by transforming latent vectors of a GAN. 
However, this approach only supports editing using a single point being dragged on the image and does not handle multiple-point constraints well.
In addition, the control is not precise, \ie, after editing, the target point is often not reached. 
\vspace{-2pt}
\paragraph{3D-aware GANs} Several methods modify the architecture of the GAN to enable 3D control~\cite{Schwarz2020NEURIPS,chan2021pi,Chan2022,gu2021stylenerf,pan2021shadegan,tewari2022d3d,sofgan,xu2022volumegan}.
Here, the model generates 3D representations that can be rendered using a physically-based analytic renderer.  
However, unlike our approach, control is limited to global pose or lighting. 

\vspace{-2pt}
\paragraph{Diffusion Models} More recently, diffusion models~\cite{sohl2015deep} have enabled image synthesis at high quality~\cite{ho2020denoising,song2020denoising,song2021scorebased}. 
These models iteratively denoise a randomly sampled noise to create a photorealistic image. 
Recent models have shown expressive image synthesis conditioned on text inputs~\cite{rombach2021highresolution,saharia2022photorealistic,ramesh2022hierarchical}. 
However, natural language does not enable fine-grained control over the spatial attributes of images, and thus, all text-conditional methods are restricted to high-level semantic editing. 
In addition, current diffusion models are slow since they require multiple denoising steps. 
While progress has been made toward efficient sampling, GANs are still significantly more efficient. 

\begin{figure*}[t]
\centering
  \includegraphics[width=16cm]{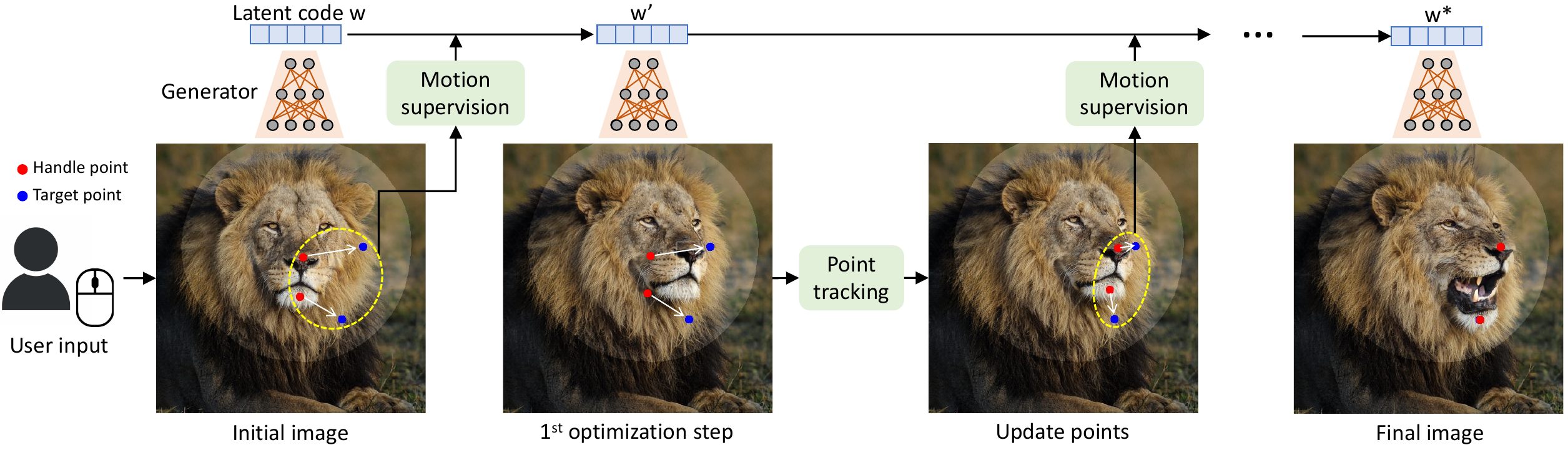}
  \vspace{-0.4cm}
  \caption{Overview of our pipeline. Given a GAN-generated image, the user only needs to set  several handle points (\textcolor{red}{red} dots), target points (\textcolor{blue}{blue} dots), and optionally a mask denoting the movable region during editing  (\textcolor{gray}{brighter} area).
  Our approach iteratively performs \textit{motion supervision} (Sec.~\ref{sec:motion_supervision}) and \textit{point tracking} (Sec.~\ref{sec:point_tracking}).
  The motion supervision step drives the handle points (red dots) to move towards the target points (blue dots) and the  point tracking step updates the handle points to track the object in the image.
  This process continues until the handle points reach their corresponding target points.
  }
  \vspace{-0.2cm}
  \label{fig:overview}
\end{figure*}

\subsection{Image Deformation}


How to deform images following users' point-drag command is a classic problem in computer graphics.
Conventional approaches~\cite{botsch2007linear} typically convert images into meshes, and then deform the mesh subject to geometric constraints such as rigidity~\cite{igarashi2005rigid,sorkine2007rigid} and Laplacian smoothness~\cite{lipman2004differential,lipman2005linear,sorkine2004laplacian}.
An earlier work has proposed the notion of shape deformation based on hand-crafted features~\cite{beier2023feature}.
However, these geometric constraints and hand-crafted features lack knowledge on the underlying structure and rigidity of the edited objects, often producing sub-optimal deformation.
Additionally, they cannot hallucinate new content when needed such as synthesizing occluded regions.
It is also shown that point-drag editing can be approximated by navigating in a video, but the need of video data limits its applicability~\cite{goldman2008video}.
In contrast, this work studies image deformation based on a strong generative image prior that captures rich information about object structure and appearance.

\subsection{Point Tracking}

To track points in videos, an obvious approach is through optical flow estimation between consecutive frames.
Optical flow estimation is a classic problem that estimates motion fields between two images.
Conventional approaches solve optimization problems with hand-crafted criteria~\cite{brox2010large,sundaram2010dense}, while deep learning-based approaches started to dominate the field in recent years due to better performance~\cite{dosovitskiy2015flownet,ilg2017flownet,teed2020raft}.
These deep learning-based approaches typically use synthetic data with ground truth optical flow to train the deep neural networks.
Among them, the most widely used method now is RAFT~\cite{teed2020raft}, which estimates optical flow via an iterative algorithm.
Recently, Harley \etal~\shortcite{harley2022particle} combines this iterative algorithm with a conventional ``particle video'' approach, giving rise to a new point tracking method named PIPs.
PIPs considers information across multiple frames and thus handles long-range tracking better than previous approaches.

In this work, we show that point tracking on GAN-generated images can be performed without using any of the aforementioned approaches or additional neural networks.
We reveal that the feature spaces of GANs are discriminative enough such that tracking can be achieved simply via feature matching.
While some previous works also leverage the discriminative feature in semantic segmentation~\cite{tritrong2021repurposing,zhang21}, we are the first to connect the point-based editing problem to the intuition of discriminative GAN features and design a concrete method.
Getting rid of additional tracking models allows our approach to run much more efficiently to support interactive editing.
Despite the simplicity of our approach, we show that it outperforms the state-of-the-art point tracking approaches including RAFT and PIPs in our experiments.

\section{Method}

This work aims to develop an interactive image manipulation method for GANs where users only need to click on the images to define some pairs of (handle point, target point) and drive the handle points to reach their corresponding target points.
Our study is based on the StyleGAN2 architecture~\cite{karras2020analyzing}.
Here we briefly introduce the basics of this architecture.

\paragraph{StyleGAN Terminology.}
In the StyleGAN2 architecture, a 512 dimensional latent code $\vz \in \mathcal{N} (0, \mI)$ is mapped to an intermediate latent code $\vw \in \mathbb{R}^{512}$ via a mapping network.
The space of $\vw$ is commonly referred to as $\mathcal{W}$.
$\vw$ is then sent to the generator $G$ to produce the output image $\mathbf{I} = G(\vw)$.
In this process, $\vw$ is copied several times and sent to different layers of the generator $G$ to control different levels of attributes.
Alternatively, one can also use different $\vw$ for different layers, in which case the input would be $\vw \in \mathbb{R}^{l \times 512} = \mathcal{W}^+$, where $l$ is the number of layers.
This less constrained $\mathcal{W}^+$ space is shown to be more expressive~\cite{abdal2019image2stylegan}.
As the generator $G$ learns a mapping from a low-dimensional latent space to a much higher dimensional image space, it can be seen as modelling an image manifold~\cite{zhu2016generative}.

\subsection{Interactive Point-based Manipulation}

An overview of our image manipulation pipeline is shown in Fig.~\ref{fig:overview}.
For any image $\mathbf{I} \in \mathbb{R}^{3 \times H \times W}$ generated by a GAN with latent code $\vw$, we allow the user to input a number of handle points $\{\vp_i=(x_{p,i}, y_{p,i}) | i=1,2,...,n\}$ and their corresponding target points $\{\vt_{i}=(x_{t,i}, y_{t,i}) | i=1,2,...,n\}$ (\ie, the corresponding target point of $\vp_i$ is $\vt_{i}$).
The goal is to move the object in the image such that the semantic positions (\eg, the nose and the jaw in Fig.~\ref{fig:overview}) of the handle points reach their corresponding target points.
We also allow the user to optionally draw a binary mask $\mathbf{M}$ denoting which region of the image is movable.

Given these user inputs, we perform image manipulation in an optimization manner.
As shown in Fig.~\ref{fig:overview}, each optimization step consists of two sub-steps, including 1) \textit{motion supervision} and 2) \textit{point tracking}.
In motion supervision, a loss that enforces handle points to move towards target points is used to optimize the latent code $\vw$.
After one optimization step, we get a new latent code $\vw'$ and a new image $\mathbf{I}'$.
The update would cause a slight movement of the object in the image.
Note that the motion supervision step only moves each handle point towards its target by a small step but the exact length of the step is unclear as it is subject to complex optimization dynamics and therefore varies for different objects and parts.
Thus, we then update the positions of the handle points $\{\vp_i\}$ to track the corresponding points on the object.
This tracking process is necessary because if the handle points (\eg, nose of the lion) are not accurately tracked, then in the next motion supervision step, wrong points (\eg, face of the lion) will be supervised, leading to undesired results.
After tracking, we repeat the above optimization step based on the new handle points and latent codes.
This optimization process continues until the handle points $\{\vp_i\}$ reach the position of the target points $\{\vt_i\}$, which usually takes 30-200 iterations in our experiments.
The user can also stop the optimization at any intermediate step.
After editing, the user can input new handle and target points and continue editing until satisfied with the results.

\begin{figure}[t]
	\centering
	\includegraphics[width=\linewidth]{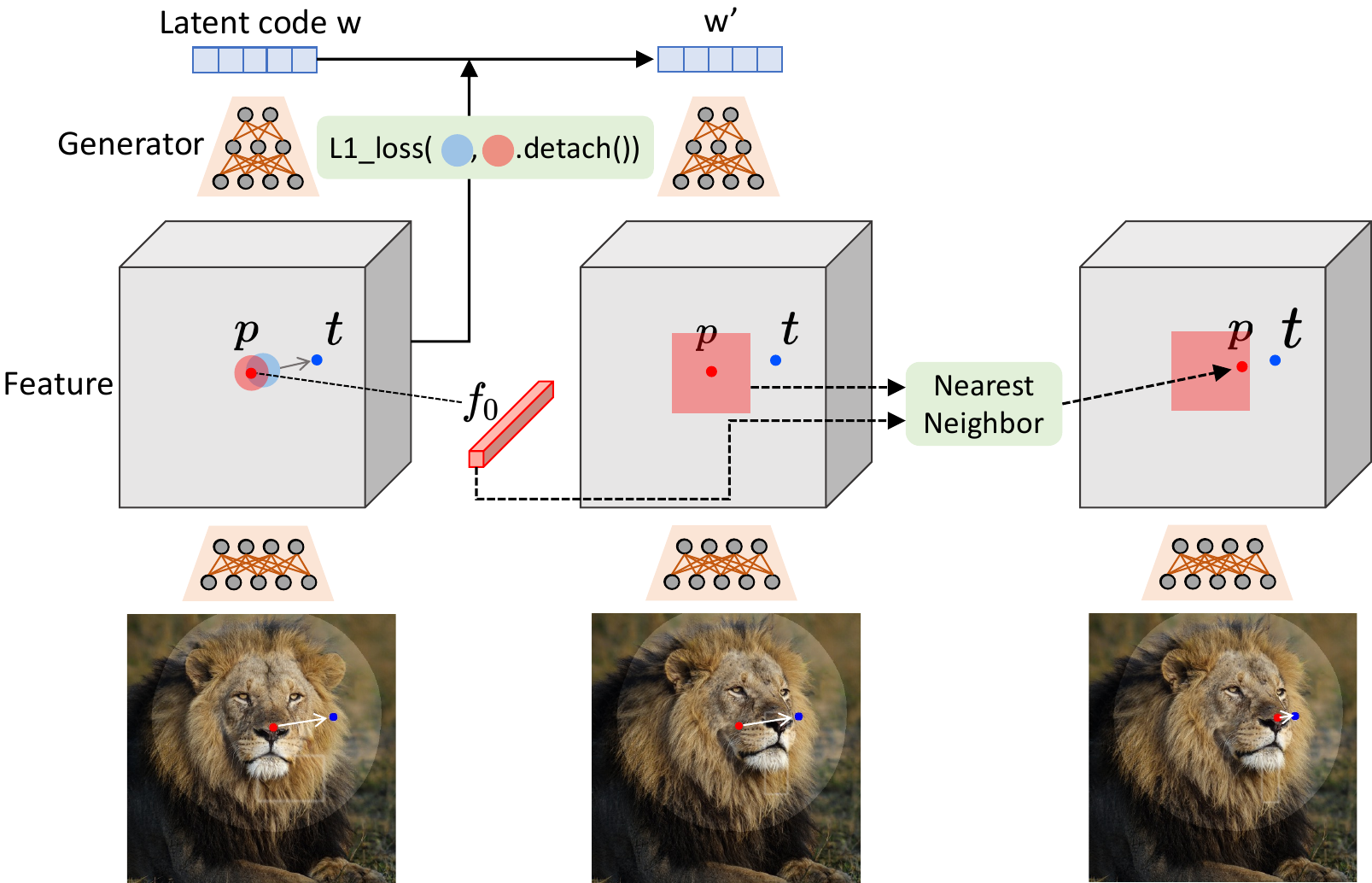}
    \vspace{-0.7cm}
	\caption{Method. Our motion supervision is achieved via a shifted patch loss on the feature maps of the generator. We perform point tracking on the same feature space via the nearest neighbor search. 
    }
    \vspace{-0.3cm}
	\label{fig:method}
\end{figure}

\begin{figure*}[t!]
	\centering
	\includegraphics[width=16cm]{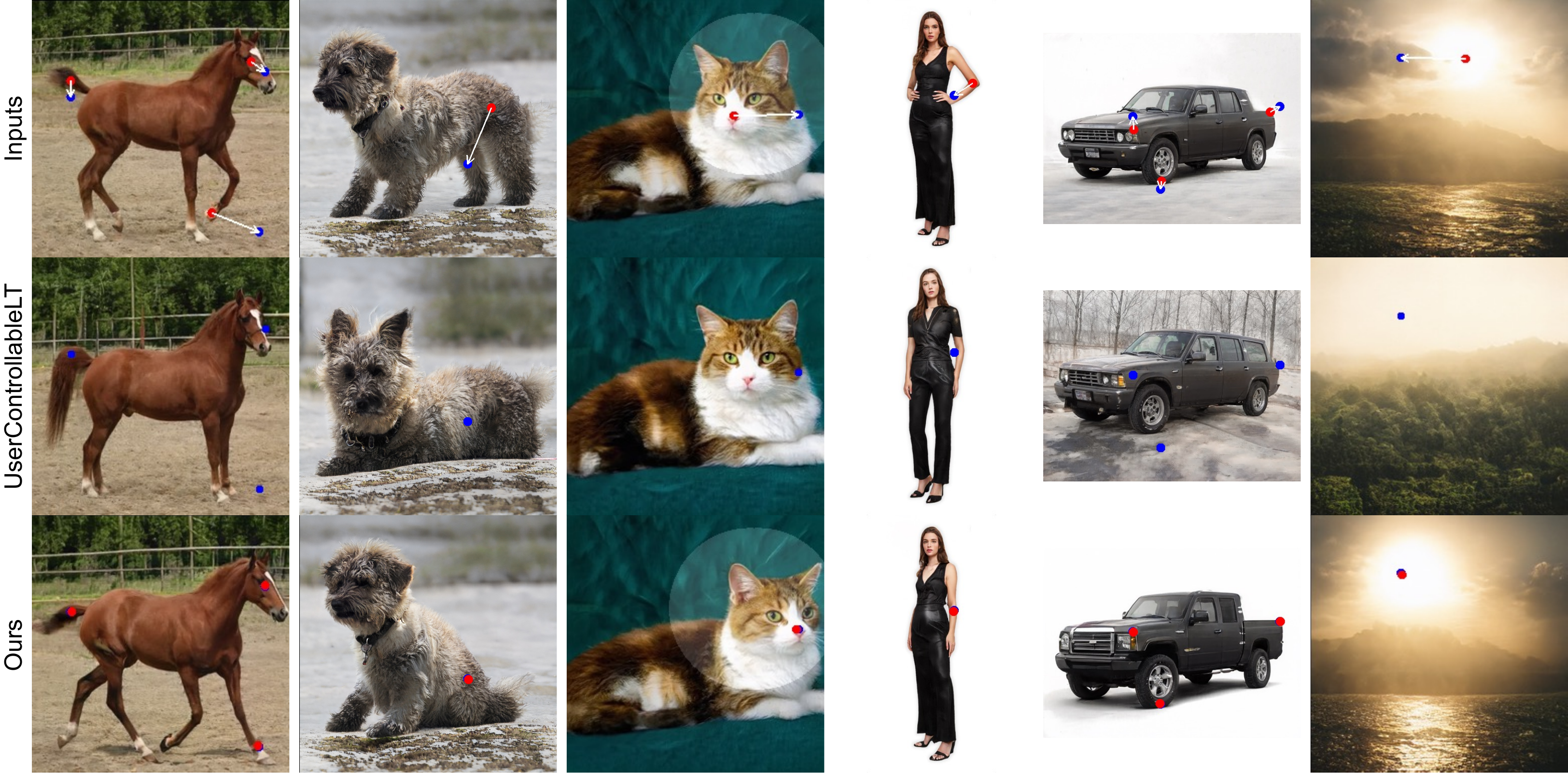}
    \vspace{-0.4cm}
	\caption{Qualitative comparison of our approach to UserControllableLT~\cite{endoPG2022} on the task of moving handle points (\textcolor{red}{red} dots) to target points (\textcolor{blue}{blue} dots). Our approach achieves more natural and superior results on various datasets. 
    More examples are provided in Fig.~\ref{fig:qualitative2}.
    }
    \vspace{-0.25cm}
	\label{fig:qualitative}
\end{figure*}

\subsection{Motion Supervision}
\label{sec:motion_supervision}
How to supervise the point motion for a GAN-generated image has not been much explored before.
In this work, we propose a motion supervision loss that does not rely on any additional neural networks.
The key idea is that the intermediate features of the generator are very discriminative such that a simple loss suffices to supervise motion. 
Specifically, we consider the feature maps $\textbf{F}$ after the 6th block of StyleGAN2, which performs the best among all features due to a good trade-off between resolution and discriminativeness.
We resize $\textbf{F}$ to have the same resolution as the final image via bilinear interpolation.
As shown in Fig.~\ref{fig:method}, to move a handle point $\vp_i$ to the target point $\vt_i$, our idea is to supervise a small patch around $\vp_i$ (red circle) to move towards $\vt_i$ by a small step (blue circle).
We use $\Omega_1(\vp_i, r_1)$ to denote the pixels whose distance to $\vp_i$ is less than $r_1$, then our motion supervision loss is:
\begin{align}
    \mathcal{L} = \sum_{i=0}^{n} \sum_{\vq_i \in \Omega_1(\vp_i, r_1)} \| \mathbf{F}(\vq_i) - \mathbf{F}(\vq_i+\vd_i) \|_1 + \lambda \| (\mathbf{F} - \mathbf{F}_0) \cdot (1-\mathbf{M}) \|_1,
    \label{eq:motion}
\end{align}
where $\mathbf{F}(\vq)$ denotes the feature values of $\mathbf{F}$ at pixel $\vq$, $\vd_i = \frac{\vt_i - \vp_i}{\|\vt_i - \vp_i\|_2}$ is a normalized vector pointing from $\vp_i$ to $\vt_i$ ($\vd_i = 0$ if $\vt_i = \vp_i$), and $\mathbf{F}_0$ is the feature maps corresponding to the initial image.
Note that the first term is summed up over all handle points $\{\vp_i\}$.
As the components of $\vq_i+\vd_i$ are not integers, we obtain $\mathbf{F}(\vq_i+\vd_i)$ via bilinear interpolation.
Importantly, when performing back-propagation using this loss, the gradient is not back-propagated through $\mathbf{F}(\vq_i)$.
This will motivate $\vp_i$ to move to $\vp_i+\vd_i$ but not vice versa.
In case the binary mask $\mathbf{M}$ is given, we keep the unmasked region fixed with a reconstruction loss shown as the second term.
At each motion supervision step, this loss is used to optimize the latent code $\vw$ for one step.
$\vw$ can be optimized either in the $\mathcal{W}$ space or in the $\mathcal{W}^+$ space, depending on whether the user wants a more constrained image manifold or not.
As $\mathcal{W}^+$ space is easier to achieve out-of-distribution manipulations (\eg, cat in Fig.~\ref{fig:W}), we use $\mathcal{W}^+$ in this work for better editability. 
In practice, we observe that the spatial attributes of the image are mainly affected by the $\vw$ for the first 6 layers while the remaining ones only affect appearance.
Thus, inspired by the style-mixing technique~\cite{karras2019style}, we only update the $\vw$ for the first 6 layers while fixing others to preserve the appearance.
This selective optimization leads to the desired slight movement of image content.

\subsection{Point Tracking}
\label{sec:point_tracking}
The previous motion supervision results in a new latent code $\vw'$, new feature maps $\mathbf{F}'$, and a new image $\mathbf{I}'$.
As the motion supervision step does not readily provide the precise new locations of the handle points, our goal here is to update each handle point $\vp_i$ such that it tracks the corresponding point on the object.
Point tracking is typically performed via optical flow estimation models or particle video approaches~\cite{harley2022particle}.
Again, these additional models can significantly harm efficiency and may suffer from accumulation error, especially in the presence of alias artifacts in GANs.
We thus present a new point tracking approach for GANs.
The insight is that the discriminative features of GANs well capture dense correspondence and thus tracking can be effectively performed via nearest neighbor search in a feature patch.
Specifically, we denote the feature of the initial handle point as $\vf_i = \textbf{F}_0 (\vp_i)$.
We denote the patch around $\vp_i$ as $\Omega_2 (\vp_i, r_2) = \{(x, y) \mid |x - x_{p,i}| < r_2, |y - y_{p,i}| < r_2 \}$.
Then the tracked point is obtained by searching for the nearest neighbor of $f_i$ in $\Omega_2 (\vp_i, r_2)$:
\begin{align}
    \vp_i := \argmin_{\vq_i \in \Omega_2 (\vp_i, r_2)} \| \mathbf{F}'(\vq_i) - \vf_i \|_1.
    \label{eq:tracking}
\end{align}
In this way, $\vp_i$ is updated to track the object.
For more than one handle point, we apply the same process for each point.
Note that here we are also considering the feature maps $\mathbf{F}'$ after the 6th block of StyleGAN2.
The feature maps have a resolution of $256 \times 256$ and are bilinear interpolated to the same size as the image if needed, which is sufficient to perform accurate tracking in our experiments.
We analyze this choice at Sec.~\ref{sec:quantitative}.

\begin{figure*}[t]
	\centering
	\includegraphics[width=\linewidth]{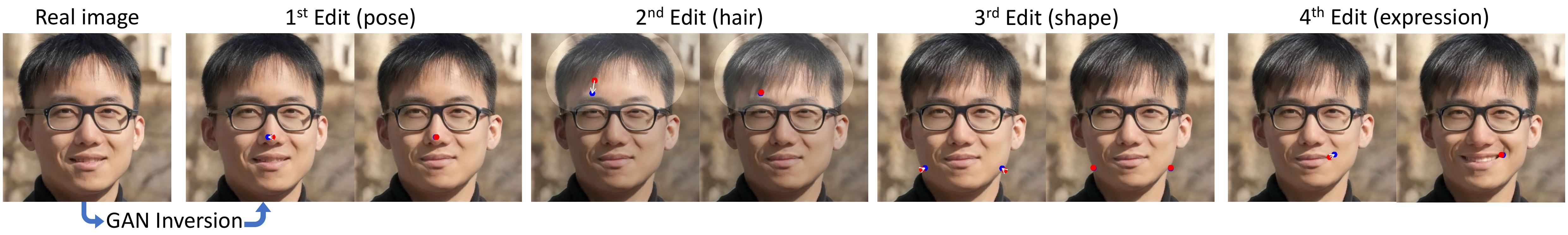}
	\vspace{-0.8cm}
	\caption{ Real image manipulation. Given a real image, we apply GAN inversion to map it to the latent space of StyleGAN, then edit the pose, hair, shape, and expression, respectively. }
	\vspace{-0.35cm}
	\label{fig:real_image}
\end{figure*}

\subsection{Implementation Details}

We implement our approach based on PyTorch~\cite{paszke2017automatic}.
We use the Adam optimizer~\cite{kingma2014adam} to optimize the latent code $\vw$ with a step size of 2e-3 for FFHQ~\cite{karras2019style}, AFHQCat~\cite{choi2020starganv2}, and LSUN Car~\cite{yu2015lsun} datasets and 1e-3 for others.
The hyper-parameters are set to be $\lambda = 20, r_1 = \round(3 / 512 \times size ), r_2 = \round(12 / 512 \times size)$, where $size$ is the resolution of the generated image.
In our implementation, we stop the optimization process when all the handle points are no more than $d$ pixel away from their corresponding target points, where $d$ is set to 1 for no more than 5 handle points and 2 otherwise.
We also develop a GUI to support interactive image manipulation.
Thanks to the computational efficiency of our approach, users only need to wait for a few 
seconds for each edit and can continue the editing until satisfied.
We highly recommend readers refer to the supplemental video for live recordings of interactive sessions.

\section{Experiments}

\paragraph{Datasets.} We evaluate our approach based on StyleGAN2~\cite{karras2020analyzing} pretrained on the following datasets (the resolution of the pretrained StyleGAN2 is shown in brackets): FFHQ (512)~\cite{karras2019style}, AFHQCat (512)~\cite{choi2020starganv2}, SHHQ (512)~\cite{fu2022styleganhuman}, LSUN Car (512)~\cite{yu2015lsun}, LSUN Cat (256)~\cite{yu2015lsun}, Landscapes HQ (256)~\cite{ALIS}, microscope (512)~\cite{Pinkney_Awesome_pretrained_StyleGAN2} and self-distilled dataset from ~\cite{mokady2022self} including Lion (512), Dog (1024), and Elephant (512).

\vspace{-3pt}
\paragraph{Baselines.} Our main baseline is UserControllableLT~\cite{endoPG2022}, which has the closest setting with our method. 
UserControllableLT does not support a mask input but allows users to define a number of fixed points.
Thus, for testing cases with a mask input, we sample a regular $16 \times 16$ grid on the image and use the points outside the mask as the fixed points to UserControllableLT.
Besides, we also compare with RAFT~\cite{teed2020raft} and PIPs~\cite{harley2022particle} for point tracking.
To do so, we create two variants of our approach where the point tracking part (Sec.\ref{sec:point_tracking}) is replaced with these two tracking methods.

\begin{figure}[t]
	\centering
	\includegraphics[width=\linewidth]{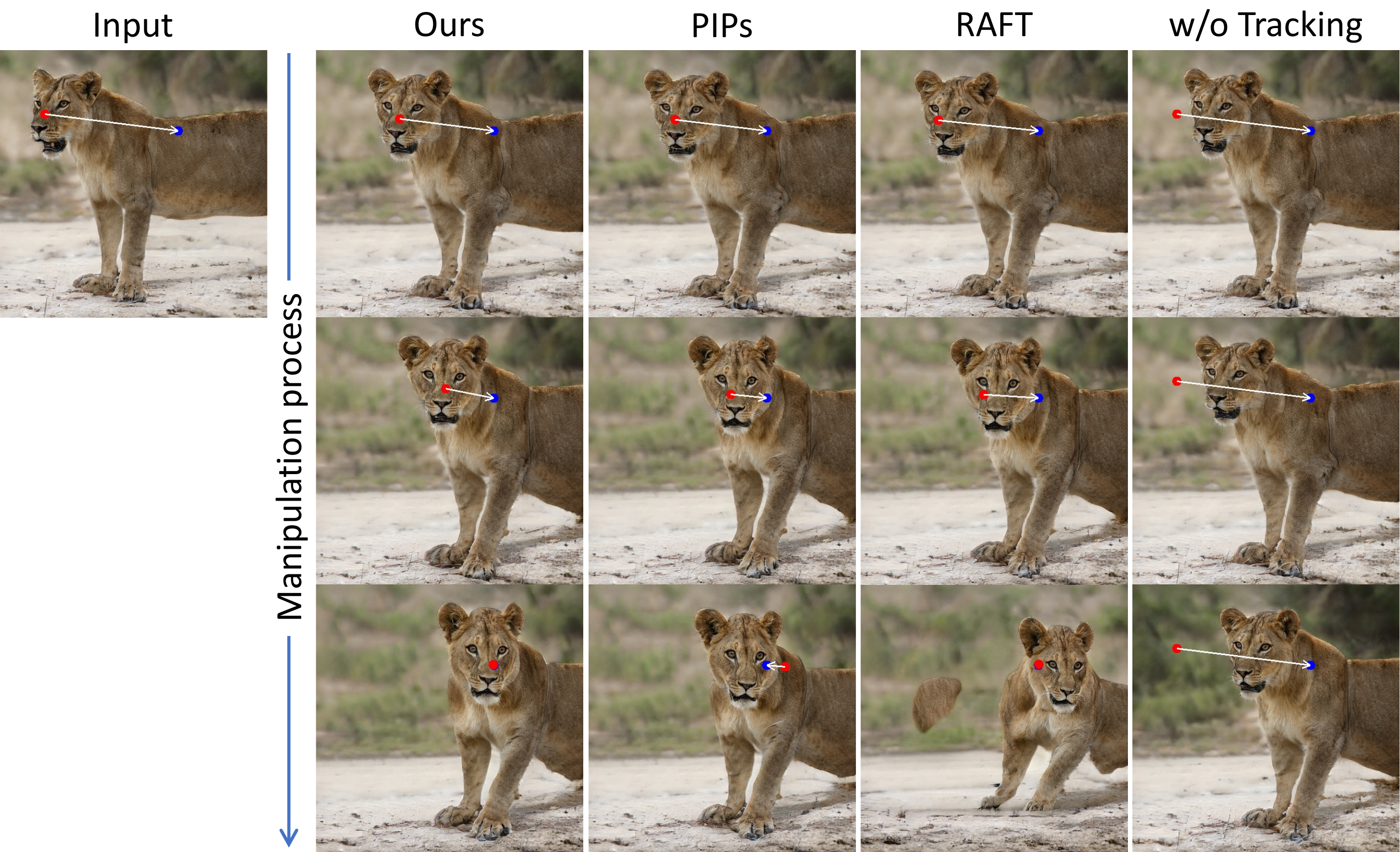}
    \vspace{-0.75cm}
	\caption{ Qualitative tracking comparison of our approach to RAFT~\cite{teed2020raft}, PIPs~\cite{harley2022particle}, and without tracking. 
    Our approach tracks the handle point more accurately than baselines, thus producing more precise editing.
    }
    \vspace{-0.35cm}
	\label{fig:tracking}
\end{figure}

\subsection{Qualitative Evaluation}

Fig.~\ref{fig:qualitative} shows the qualitative comparison between our method and UserControllableLT.
We show the image manipulation results for several different object categories and user inputs.
Our approach accurately moves the handle points to reach the target points, achieving diverse and natural manipulation effects such as changing the pose of animals, the shape of a car, and the layout of a landscape.
In contrast, UserControllableLT cannot faithfully move the handle points to the targets and often leads to undesired changes in the images, \eg, the clothes of the human and the background of the car.
It also does not keep the unmasked region fixed as well as ours, as shown in the cat images.
We show more comparisons in Fig.~\ref{fig:qualitative2}.

A comparison between our approach with PIPs and RAFT is provided in Fig.~\ref{fig:tracking}.
Our approach accurately tracks the handle point above the nose of the lion, thus successfully driving it to the target position.
In PIPs and RAFT, the tracked point starts to deviate from the nose during the manipulation process.
Consequently, they move the wrong part to the target position.
When no tracking is performed, the fixed handle point soon starts to drive another part of the image (\eg, background) after a few steps and never knows when to stop, which fails to achieve the editing goal.

\vspace{-3pt}
\paragraph{Real image editing.}
Using GAN inversion techniques that embed a real image in the latent space of StyleGAN, we can also apply our approach to manipulate real images.
Fig.~\ref{fig:real_image} shows an example, where we apply PTI inversion~\cite{roich2022pivotal} to the real image and then perform a series of manipulations to edit the pose, hair, shape, and expression of the face in the image.
We show more real image editing examples in Fig.~\ref{fig:real}.

\subsection{Quantitative Evaluation}
\label{sec:quantitative}

We quantitatively evaluate our method under two settings, including face landmark manipulation and paired image reconstruction.

\begin{figure}[t]
	\centering
    \vspace{-0.2cm}
	\includegraphics[width=\linewidth]{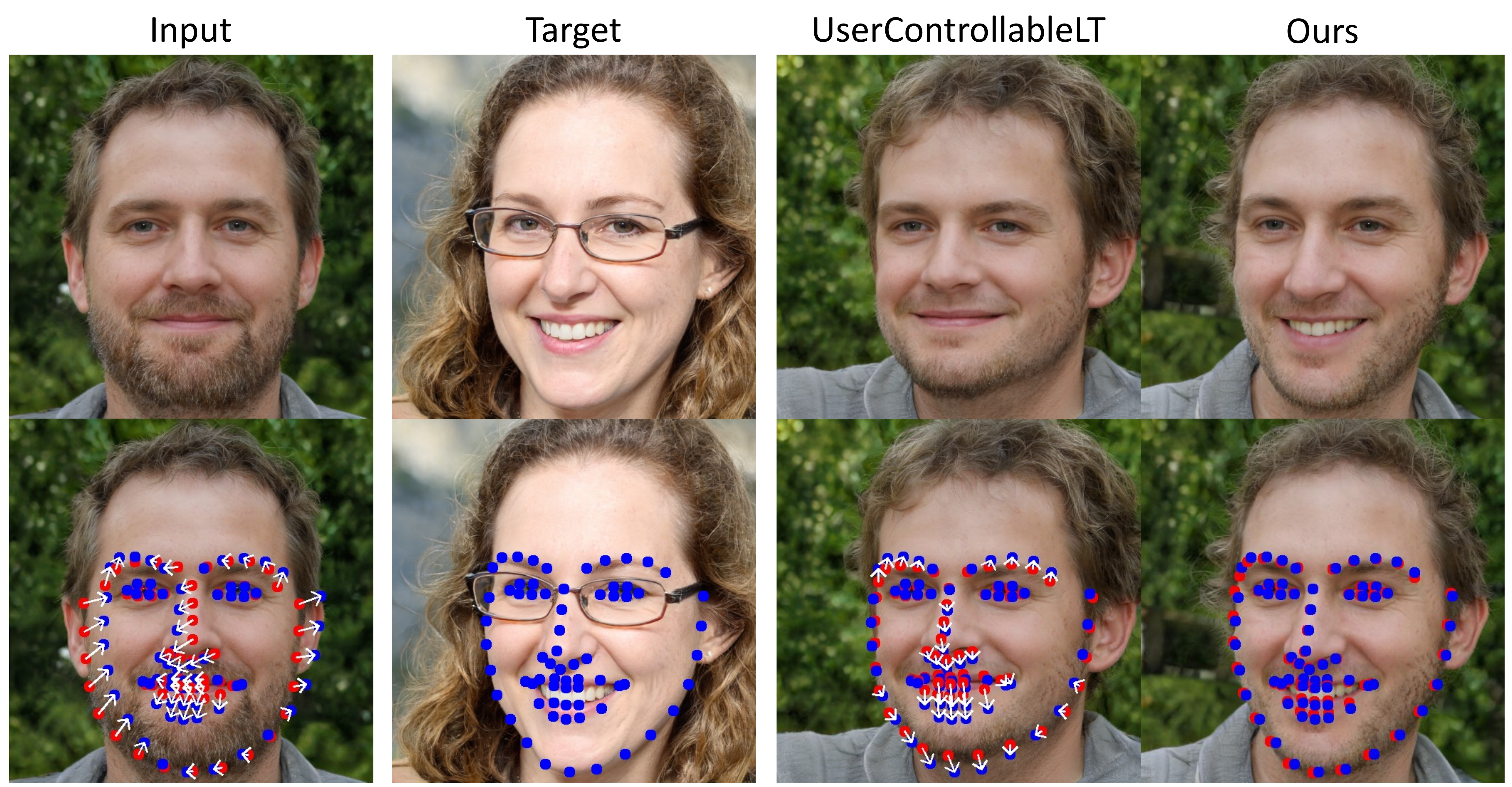}
    \vspace{-0.75cm}
	\caption{ Face landmark manipulation. Compared to UserControllableLT~\cite{endoPG2022}, our method can manipulate the landmarks detected from the input image to match the landmarks detected from the target image with less matching error.  }
    \vspace{-0.25cm}
	\label{fig:landmarks}
\end{figure}

\begin{table}
\centering
\caption{Quantitative evaluation on face keypoint manipulation. We compute the mean distance between edited points and target points. The FID and Time are reported based on the `1 point' setting.}
\vspace{-0.3cm}
\resizebox{0.48\textwidth}{!}{
\begin{tabular}{lccccc}
\hline
Method             & 1 point & 5 points & 68 points & FID & Time (s) \\ \hline
No edit            &   12.93      &  11.66   &  16.02  &  -  &  -  \\
UserControllableLT &  11.64   &  10.41    &  10.15  &  25.32  &  0.03    \\
Ours w. RAFT tracking      &   13.43      &  13.59    &  15.92    & 51.37  &   15.4   \\
Ours w. PIPs tracking      &  2.98       &  4.83    &   5.30    &  31.87  &   6.6   \\
Ours               &  \textbf{2.44}   &  \textbf{3.18} &   \textbf{4.73}    &  \textbf{9.28}   &  2.0  \\ \hline
\vspace{-0.7cm}
\end{tabular}
\label{tab:quant1}
}
\end{table}

\begin{table*}[t]
\centering
	\begin{minipage}{0.6\linewidth}
\caption{Quantitative evaluation on paired image reconstruction. We follow the evaluation in~\cite{endoPG2022} and report MSE $(\times 10^2)$$\downarrow$ and LPIPS $(\times 10)$$\downarrow$ scores. }
\vspace{-0.3cm}
\centering
\resizebox{11cm}{!}{
\begin{tabular}{lcccccccc}
\hline
Dataset            & \multicolumn{2}{c}{Lion} & \multicolumn{2}{c}{LSUN Cat} & \multicolumn{2}{c}{Dog} & \multicolumn{2}{c}{LSUN Car} \\
Metric             & MSE    & LPIPS    & MSE    & LPIPS   & MSE    & LPIPS    & MSE   & LPIPS      \\ \hline
UserControllableLT & 1.82   &  1.14 & 1.25   & 0.87        & 1.23   &  0.92   & 1.98   &  0.85     \\
Ours w. RAFT tracking  & 1.09 & 0.99 & 1.84 & 1.15 & 0.91 & 0.76 & 2.37 & 0.94 \\
Ours w. PIPs tracking  & 0.80 & 0.82 & 1.11 & 0.85 & 0.78 & 0.63 & 1.81 & 0.79 \\
Ours               &  \textbf{0.66}  &  \textbf{0.72} & \textbf{1.04}   & \textbf{0.82}       &   \textbf{0.48}  &  \textbf{0.44}   &  \textbf{1.67}  &  \textbf{0.74}         \\ \hline
\end{tabular}
}
\label{tab:quant2}
    \end{minipage}
    \hfill
	\begin{minipage}{0.38\linewidth}
 \caption{Effects of which feature to use. x+y means the concatenation of two features. We report the performance (MD) of face landmark manipulation (1 point).}
\vspace{-0.4cm}
\centering
\resizebox{6.8cm}{!}{
\begin{tabular}{lcccccc}
\hline
Block No. & 4    & 5    & 6    & 7    & 5+6 & 6+7    \\ \hline
Motion sup. & 2.73 & 2.50 & \textbf{2.44} & 2.51 & 2.47 & 2.45 \\
Tracking & 3.61 & 2.55 & \textbf{2.44} & 2.58 & 2.47 & 2.45 \\ \hline
\end{tabular}
}
\label{tab:ablation_feature}

\caption{Effects of $r_1$.}
\vspace{-0.4cm}
\centering
\begin{tabular}{lccccc}
\hline
$r_1$ & 1    & 2    & 3    & 4    & 5    \\ \hline
MD & 2.49 & 2.51 & \textbf{2.44} & 2.45 & 2.46 \\ \hline
\end{tabular}
\label{tab:ablation_r1}
\vspace{-0.25cm}

    \end{minipage}
\end{table*}


\vspace{-3pt}
\paragraph{Face landmark manipulation.}
Since face landmark detection is very reliable using an off-the-shelf tool~\cite{dlib09}, we use its prediction as ground truth landmarks.
Specifically, we randomly generate two face images using the StyleGAN trained on FFHQ and detect their landmarks.
The goal is to manipulate the landmarks of the first image to match the landmarks of the second image.
After manipulation, we detect the landmarks of the final image and compute the mean distance (MD) to the target landmarks.
The results are averaged over 1000 tests.
The same set of test samples is used to evaluate all methods.
In this way, the final MD score reflects how well the method can move the landmarks to the target positions.
We perform the evaluation under 3 settings with different numbers of landmarks including 1, 5, and 68 to show the robustness of our approach under different numbers of handle points. 
We also report the FID score between the edited images and the initial images as an indication of image quality.
In our approach and its variants, the maximum optimization step is set to 300.

The results are provided in Table~\ref{tab:quant1}.
Our approach significantly outperforms UserControllableLT under different numbers of points.
A qualitative comparison is shown in Fig.~\ref{fig:landmarks}, where our method opens the mouth and adjusts the shape of the jaw to match the target face while UserControllableLT fails to do so.
Furthermore, our approach preserves better image quality as indicated by the FID scores.
Thanks to a better tracking capability, we also achieve more accurate manipulation than RAFT and PIPs.
Inaccurate tracking also leads to excessive manipulation, which deteriorates the image quality as shown in FID scores.
Although UserControllableLT is faster, our approach largely pushes the upper bound of this task, achieving much more faithful manipulation while maintaining a comfortable running time for users.

\vspace{-4pt}
\paragraph{Paired image reconstruction.}
In this evaluation, we follow the same setting as UserControllableLT~\cite{endoPG2022}.
Specifically, we sample a latent code $\vw_1$ and randomly perturb it to get $\vw_2$ in the same way as in~\cite{endoPG2022}.
Let $\mathbf{I}_1$ and $\mathbf{I}_2$ be the StyleGAN images generated from the two latent codes.
We then compute the optical flow between $\mathbf{I}_1$ and $\mathbf{I}_2$ and randomly sample 32 pixels from the flow field as the user input $\mathcal{U}$.
The goal is to reconstruct $\mathbf{I}_2$ from $\mathbf{I}_1$ and $\mathcal{U}$.
We report MSE and LPIPS~\cite{zhang2018perceptual} and average the results over 1000 samples.
The maximum optimization step is set to 100 in our approach and its variants.
As shown in Table \ref{tab:quant2}, our approach outperforms all the baselines in different object categories, which is consistent with previous results.

\vspace{-4pt}
\paragraph{Ablation Study.}


Here we study the effects of which feature to use in motion supervision and point tracking. 
We report the performance (MD) of face landmark manipulation using different features.
As Table~\ref{tab:ablation_feature} shows, in both motion supervision and point tracking, the feature maps after the 6th block of StyleGAN perform the best, showing the best balance between resolution and discriminativeness.
We also provide the effects of $r_1$ in Table~\ref{tab:ablation_r1}.
It can be observed that the performance is not very sensitive to the choice of $r_1$, and $r_1 = 3$ performs slightly better.

\subsection{Discussions}

\begin{figure}[t]
	\centering
	\vspace{-0.45cm}
	\includegraphics[width=\linewidth]{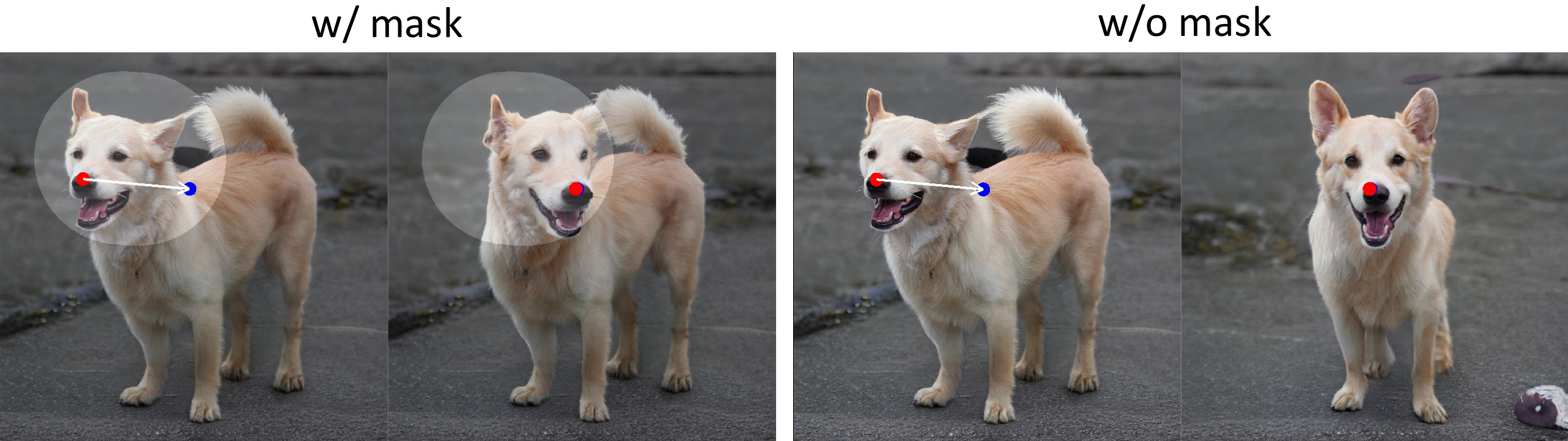}
	\vspace{-0.7cm}
	\caption{ Effects of the mask. Our approach allows masking the movable region. After masking the head region of the dog, the rest part would be almost unchanged. }
	\vspace{-0.2cm}
	\label{fig:mask}
\end{figure}

\paragraph{Effects of mask.}
Our approach allows users to input a binary mask denoting the movable region.
We show its effects in Fig.~\ref{fig:mask}.
When a mask over the head of the dog is given, the other regions are almost fixed and only the head moves.
Without the mask, the manipulation moves the whole dog's body.
This also shows that point-based manipulation often has multiple possible solutions and the GAN will tend to find the closest solution in the image manifold learned from the training data.
The mask function can help to reduce ambiguity and keep certain regions fixed.

\begin{figure}[t]
	\centering
    \vspace{-0.13cm}
	\includegraphics[width=\linewidth]{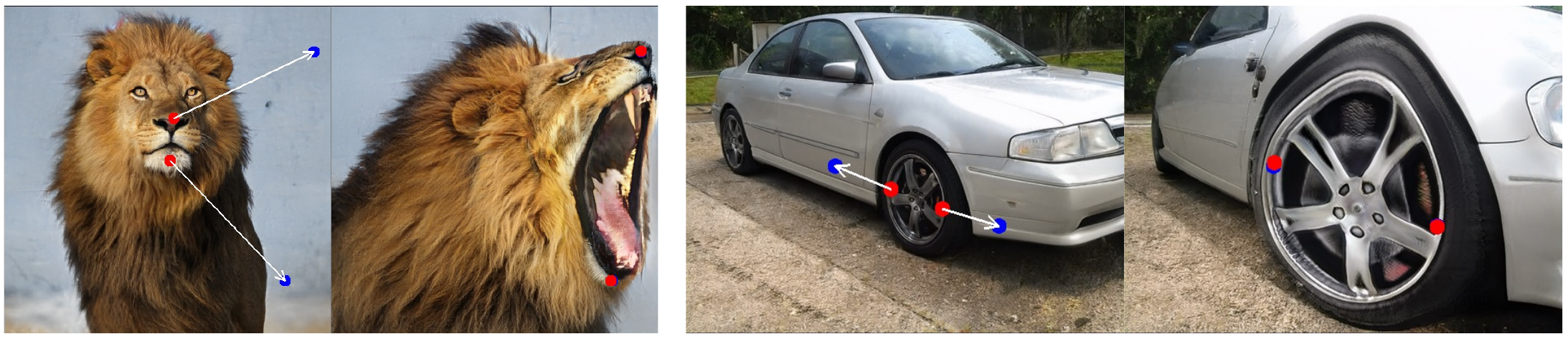}
    \vspace{-0.7cm}
	\caption{ Out-of-distribution manipulations. Our approach has extrapolation capability for creating images out of the training image distribution, for example, an extremely opened mouth and a greatly enlarged wheel. }
    \vspace{-0.2cm}
	\label{fig:outofdir}
\end{figure}

\begin{figure*}[h]
	\centering
	\includegraphics[width=16.0cm]{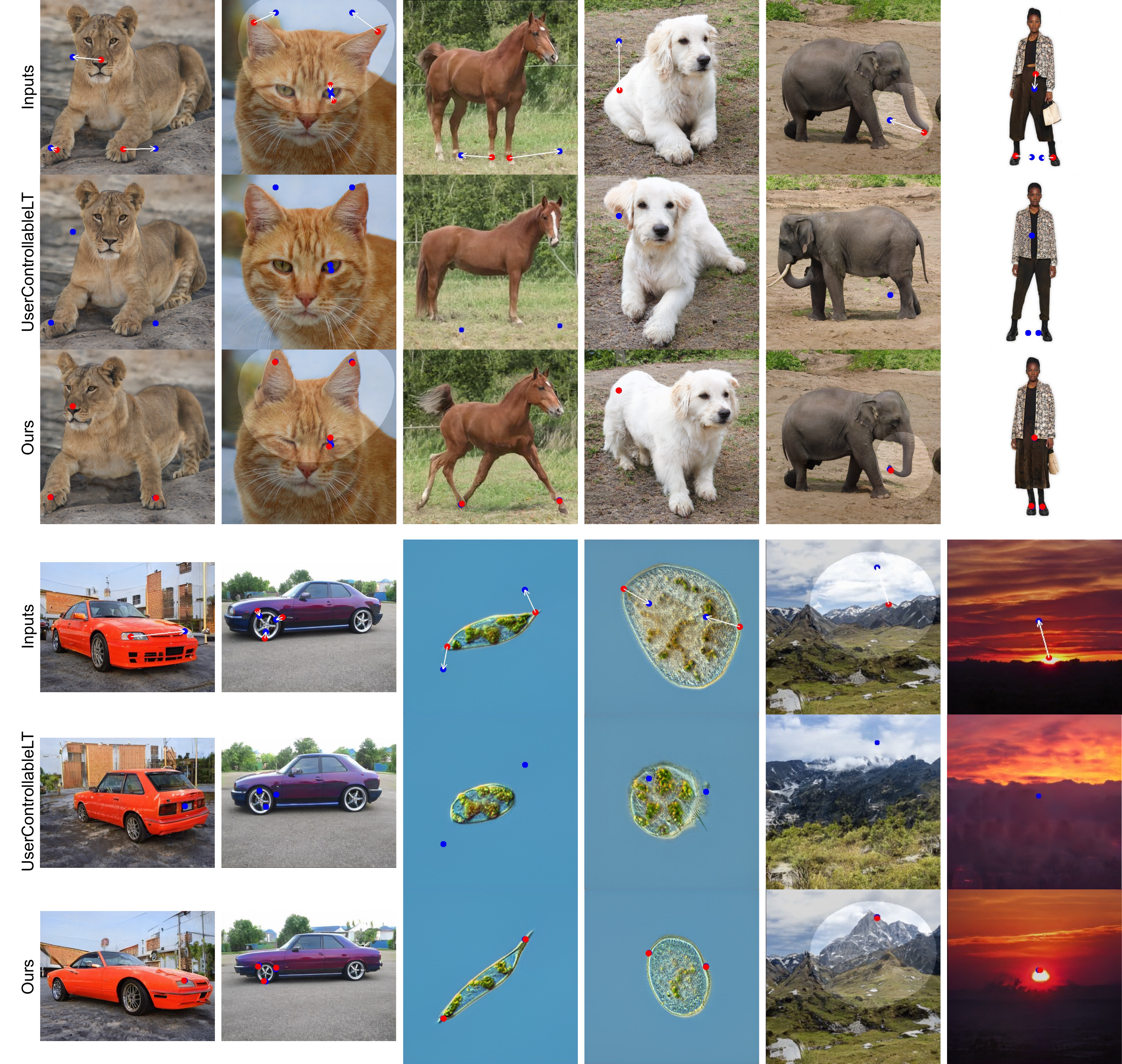}
    \vspace{-0.2cm}
	\caption{Qualitative comparison. This is an extension of Fig.~\ref{fig:qualitative}. }
	\label{fig:qualitative2}
\end{figure*}

\vspace{-4pt}
\paragraph{Out-of-distribution manipulation.}
So far, the point-based manipulations we have shown are "in-distribution" manipulations, \ie, it is possible to satisfy the manipulation requirements with a natural image inside the image distribution of the training dataset.
Here we showcase some out-of-distribution manipulations in Fig.~\ref{fig:outofdir}.
It can be seen that our approach has some extrapolation capability, creating images outside the training image distribution, \eg, an extremely opened mouth and a large wheel. 
In some cases, users may want to always keep the image in the training distribution and prevent it from reaching such out-of-distribution manipulations.
A potential way to achieve this is to add additional regularization to the latent code $\vw$, which is not the main focus of this paper.

\vspace{-4pt}
\paragraph{Limitations.}
Despite some extrapolation capability, our editing quality is still affected by the diversity of training data.
As exemplified in Fig.~\ref{fig:limitation} (a), creating a human pose that deviates from the training distribution can lead to artifacts.
Besides, handle points in texture-less regions sometimes suffer from more drift in tracking, as shown in Fig.~\ref{fig:limitation} (b)(c).
We thus suggest picking texture-rich handle points if possible.
\vspace{-4pt}
\paragraph{Social impacts.}
As our method can change the spatial attributes of images, it could be misused to create images of a real person with a fake pose, expression, or shape.
Thus, any application or research that uses our approach has to strictly respect personality rights and privacy regulations.

\section{Conclusion}

We have presented \emph{DragGAN}, an interactive approach for intuitive point-based image editing.
Our method leverages a pre-trained GAN to synthesize images that not only precisely follow user input, but also stay on the manifold of realistic images.
In contrast to many previous approaches, we present a general framework by not relying on domain-specific modeling or auxiliary networks.
This is achieved using two novel ingredients:
An optimization of latent codes that incrementally moves multiple handle points towards their target locations, and a point tracking procedure to faithfully trace the trajectory of the handle points.
Both components utilize the discriminative quality of intermediate feature maps of the GAN to yield pixel-precise image deformations and interactive performance.
We have demonstrated that our approach outperforms the state of the art in GAN-based manipulation and opens new directions for powerful image editing using generative priors.
As for future work, we plan to extend point-based editing to 3D generative models.

\begin{acks}
Christian Theobalt was supported by ERC Consolidator Grant 4DReply (770784). Lingjie Liu was supported by Lise Meitner Postdoctoral Fellowship. This project was also supported by Saarbrücken Research Center for Visual Computing, Interaction and AI.
\end{acks}

\bibliographystyle{ACM-Reference-Format}
\bibliography{paper}


\begin{thebibliography}{68}


\ifx \showCODEN    \undefined \def \showCODEN     #1{\unskip}     \fi
\ifx \showDOI      \undefined \def \showDOI       #1{#1}\fi
\ifx \showISBNx    \undefined \def \showISBNx     #1{\unskip}     \fi
\ifx \showISBNxiii \undefined \def \showISBNxiii  #1{\unskip}     \fi
\ifx \showISSN     \undefined \def \showISSN      #1{\unskip}     \fi
\ifx \showLCCN     \undefined \def \showLCCN      #1{\unskip}     \fi
\ifx \shownote     \undefined \def \shownote      #1{#1}          \fi
\ifx \showarticletitle \undefined \def \showarticletitle #1{#1}   \fi
\ifx \showURL      \undefined \def \showURL       {\relax}        \fi
\providecommand\bibfield[2]{#2}
\providecommand\bibinfo[2]{#2}
\providecommand\natexlab[1]{#1}
\providecommand\showeprint[2][]{arXiv:#2}

\bibitem[Abdal et~al\mbox{.}(2019)]%
        {abdal2019image2stylegan}
\bibfield{author}{\bibinfo{person}{Rameen Abdal}, \bibinfo{person}{Yipeng Qin},
  {and} \bibinfo{person}{Peter Wonka}.} \bibinfo{year}{2019}\natexlab{}.
\newblock \showarticletitle{Image2stylegan: How to embed images into the
  stylegan latent space?}. In \bibinfo{booktitle}{\emph{ICCV}}.
\newblock


\bibitem[Abdal et~al\mbox{.}(2021)]%
        {abdal2021styleflow}
\bibfield{author}{\bibinfo{person}{Rameen Abdal}, \bibinfo{person}{Peihao Zhu},
  \bibinfo{person}{Niloy~J Mitra}, {and} \bibinfo{person}{Peter Wonka}.}
  \bibinfo{year}{2021}\natexlab{}.
\newblock \showarticletitle{Styleflow: Attribute-conditioned exploration of
  stylegan-generated images using conditional continuous normalizing flows}.
\newblock \bibinfo{journal}{\emph{ACM Transactions on Graphics (ToG)}}
  \bibinfo{volume}{40}, \bibinfo{number}{3} (\bibinfo{year}{2021}),
  \bibinfo{pages}{1--21}.
\newblock


\bibitem[Beier and Neely(2023)]%
        {beier2023feature}
\bibfield{author}{\bibinfo{person}{Thaddeus Beier} {and} \bibinfo{person}{Shawn
  Neely}.} \bibinfo{year}{2023}\natexlab{}.
\newblock \showarticletitle{Feature-based image metamorphosis}.
\newblock In \bibinfo{booktitle}{\emph{Seminal Graphics Papers: Pushing the
  Boundaries, Volume 2}}. \bibinfo{pages}{529--536}.
\newblock


\bibitem[Botsch and Sorkine(2007)]%
        {botsch2007linear}
\bibfield{author}{\bibinfo{person}{Mario Botsch} {and} \bibinfo{person}{Olga
  Sorkine}.} \bibinfo{year}{2007}\natexlab{}.
\newblock \showarticletitle{On linear variational surface deformation methods}.
\newblock \bibinfo{journal}{\emph{IEEE transactions on visualization and
  computer graphics}} \bibinfo{volume}{14}, \bibinfo{number}{1}
  (\bibinfo{year}{2007}), \bibinfo{pages}{213--230}.
\newblock


\bibitem[Brox and Malik(2010)]%
        {brox2010large}
\bibfield{author}{\bibinfo{person}{Thomas Brox} {and} \bibinfo{person}{Jitendra
  Malik}.} \bibinfo{year}{2010}\natexlab{}.
\newblock \showarticletitle{Large displacement optical flow: descriptor
  matching in variational motion estimation}.
\newblock \bibinfo{journal}{\emph{IEEE transactions on pattern analysis and
  machine intelligence}} \bibinfo{volume}{33}, \bibinfo{number}{3}
  (\bibinfo{year}{2010}), \bibinfo{pages}{500--513}.
\newblock


\bibitem[Chan et~al\mbox{.}(2022)]%
        {Chan2022}
\bibfield{author}{\bibinfo{person}{Eric~R. Chan}, \bibinfo{person}{Connor~Z.
  Lin}, \bibinfo{person}{Matthew~A. Chan}, \bibinfo{person}{Koki Nagano},
  \bibinfo{person}{Boxiao Pan}, \bibinfo{person}{Shalini~De Mello},
  \bibinfo{person}{Orazio Gallo}, \bibinfo{person}{Leonidas Guibas},
  \bibinfo{person}{Jonathan Tremblay}, \bibinfo{person}{Sameh Khamis},
  \bibinfo{person}{Tero Karras}, {and} \bibinfo{person}{Gordon Wetzstein}.}
  \bibinfo{year}{2022}\natexlab{}.
\newblock \showarticletitle{Efficient Geometry-aware {3D} Generative
  Adversarial Networks}. In \bibinfo{booktitle}{\emph{CVPR}}.
\newblock


\bibitem[Chan et~al\mbox{.}(2021)]%
        {chan2021pi}
\bibfield{author}{\bibinfo{person}{Eric~R Chan}, \bibinfo{person}{Marco
  Monteiro}, \bibinfo{person}{Petr Kellnhofer}, \bibinfo{person}{Jiajun Wu},
  {and} \bibinfo{person}{Gordon Wetzstein}.} \bibinfo{year}{2021}\natexlab{}.
\newblock \showarticletitle{pi-gan: Periodic implicit generative adversarial
  networks for 3d-aware image synthesis}. In \bibinfo{booktitle}{\emph{CVPR}}.
\newblock


\bibitem[Chen et~al\mbox{.}(2022)]%
        {sofgan}
\bibfield{author}{\bibinfo{person}{Anpei Chen}, \bibinfo{person}{Ruiyang Liu},
  \bibinfo{person}{Ling Xie}, \bibinfo{person}{Zhang Chen},
  \bibinfo{person}{Hao Su}, {and} \bibinfo{person}{Jingyi Yu}.}
  \bibinfo{year}{2022}\natexlab{}.
\newblock \showarticletitle{Sofgan: A portrait image generator with dynamic
  styling}.
\newblock \bibinfo{journal}{\emph{ACM Transactions on Graphics (TOG)}}
  \bibinfo{volume}{41}, \bibinfo{number}{1} (\bibinfo{year}{2022}),
  \bibinfo{pages}{1--26}.
\newblock


\bibitem[Choi et~al\mbox{.}(2020)]%
        {choi2020starganv2}
\bibfield{author}{\bibinfo{person}{Yunjey Choi}, \bibinfo{person}{Youngjung
  Uh}, \bibinfo{person}{Jaejun Yoo}, {and} \bibinfo{person}{Jung-Woo Ha}.}
  \bibinfo{year}{2020}\natexlab{}.
\newblock \showarticletitle{StarGAN v2: Diverse Image Synthesis for Multiple
  Domains}. In \bibinfo{booktitle}{\emph{CVPR}}.
\newblock


\bibitem[Collins et~al\mbox{.}(2020)]%
        {collins2020editing}
\bibfield{author}{\bibinfo{person}{Edo Collins}, \bibinfo{person}{Raja Bala},
  \bibinfo{person}{Bob Price}, {and} \bibinfo{person}{Sabine Susstrunk}.}
  \bibinfo{year}{2020}\natexlab{}.
\newblock \showarticletitle{Editing in style: Uncovering the local semantics of
  gans}. In \bibinfo{booktitle}{\emph{CVPR}}. \bibinfo{pages}{5771--5780}.
\newblock


\bibitem[Creswell et~al\mbox{.}(2018)]%
        {creswell2018generative}
\bibfield{author}{\bibinfo{person}{Antonia Creswell}, \bibinfo{person}{Tom
  White}, \bibinfo{person}{Vincent Dumoulin}, \bibinfo{person}{Kai
  Arulkumaran}, \bibinfo{person}{Biswa Sengupta}, {and} \bibinfo{person}{Anil~A
  Bharath}.} \bibinfo{year}{2018}\natexlab{}.
\newblock \showarticletitle{Generative adversarial networks: An overview}.
\newblock \bibinfo{journal}{\emph{IEEE signal processing magazine}}
  \bibinfo{volume}{35}, \bibinfo{number}{1} (\bibinfo{year}{2018}),
  \bibinfo{pages}{53--65}.
\newblock


\bibitem[Deng et~al\mbox{.}(2020)]%
        {deng2020disentangled}
\bibfield{author}{\bibinfo{person}{Yu Deng}, \bibinfo{person}{Jiaolong Yang},
  \bibinfo{person}{Dong Chen}, \bibinfo{person}{Fang Wen}, {and}
  \bibinfo{person}{Xin Tong}.} \bibinfo{year}{2020}\natexlab{}.
\newblock \showarticletitle{Disentangled and Controllable Face Image Generation
  via 3D Imitative-Contrastive Learning}. In \bibinfo{booktitle}{\emph{CVPR}}.
\newblock


\bibitem[Dosovitskiy et~al\mbox{.}(2015)]%
        {dosovitskiy2015flownet}
\bibfield{author}{\bibinfo{person}{Alexey Dosovitskiy},
  \bibinfo{person}{Philipp Fischer}, \bibinfo{person}{Eddy Ilg},
  \bibinfo{person}{Philip Hausser}, \bibinfo{person}{Caner Hazirbas},
  \bibinfo{person}{Vladimir Golkov}, \bibinfo{person}{Patrick Van Der~Smagt},
  \bibinfo{person}{Daniel Cremers}, {and} \bibinfo{person}{Thomas Brox}.}
  \bibinfo{year}{2015}\natexlab{}.
\newblock \showarticletitle{Flownet: Learning optical flow with convolutional
  networks}. In \bibinfo{booktitle}{\emph{ICCV}}.
\newblock


\bibitem[Endo(2022)]%
        {endoPG2022}
\bibfield{author}{\bibinfo{person}{Yuki Endo}.}
  \bibinfo{year}{2022}\natexlab{}.
\newblock \showarticletitle{User-Controllable Latent Transformer for StyleGAN
  Image Layout Editing}.
\newblock \bibinfo{journal}{\emph{Computer Graphics Forum}}
  \bibinfo{volume}{41}, \bibinfo{number}{7} (\bibinfo{year}{2022}),
  \bibinfo{pages}{395--406}.
\newblock
\urldef\tempurl%
\url{https://doi.org/10.1111/cgf.14686}
\showDOI{\tempurl}


\bibitem[Epstein et~al\mbox{.}(2022)]%
        {epstein2022blobgan}
\bibfield{author}{\bibinfo{person}{Dave Epstein}, \bibinfo{person}{Taesung
  Park}, \bibinfo{person}{Richard Zhang}, \bibinfo{person}{Eli Shechtman},
  {and} \bibinfo{person}{Alexei~A Efros}.} \bibinfo{year}{2022}\natexlab{}.
\newblock \showarticletitle{Blobgan: Spatially disentangled scene
  representations}. In \bibinfo{booktitle}{\emph{ECCV}}.
  \bibinfo{pages}{616--635}.
\newblock


\bibitem[Fu et~al\mbox{.}(2022)]%
        {fu2022styleganhuman}
\bibfield{author}{\bibinfo{person}{Jianglin Fu}, \bibinfo{person}{Shikai Li},
  \bibinfo{person}{Yuming Jiang}, \bibinfo{person}{Kwan-Yee Lin},
  \bibinfo{person}{Chen Qian}, \bibinfo{person}{Chen-Change Loy},
  \bibinfo{person}{Wayne Wu}, {and} \bibinfo{person}{Ziwei Liu}.}
  \bibinfo{year}{2022}\natexlab{}.
\newblock \showarticletitle{StyleGAN-Human: A Data-Centric Odyssey of Human
  Generation}. In \bibinfo{booktitle}{\emph{ECCV}}.
\newblock


\bibitem[Ghosh et~al\mbox{.}(2020)]%
        {ghosh2020gif}
\bibfield{author}{\bibinfo{person}{Partha Ghosh}, \bibinfo{person}{Pravir~Singh
  Gupta}, \bibinfo{person}{Roy Uziel}, \bibinfo{person}{Anurag Ranjan},
  \bibinfo{person}{Michael~J Black}, {and} \bibinfo{person}{Timo Bolkart}.}
  \bibinfo{year}{2020}\natexlab{}.
\newblock \showarticletitle{GIF: Generative interpretable faces}. In
  \bibinfo{booktitle}{\emph{International Conference on 3D Vision (3DV)}}.
\newblock


\bibitem[Goldman et~al\mbox{.}(2008)]%
        {goldman2008video}
\bibfield{author}{\bibinfo{person}{Dan~B Goldman}, \bibinfo{person}{Chris
  Gonterman}, \bibinfo{person}{Brian Curless}, \bibinfo{person}{David Salesin},
  {and} \bibinfo{person}{Steven~M Seitz}.} \bibinfo{year}{2008}\natexlab{}.
\newblock \showarticletitle{Video object annotation, navigation, and
  composition}. In \bibinfo{booktitle}{\emph{Proceedings of the 21st annual ACM
  symposium on User interface software and technology}}.
  \bibinfo{pages}{3--12}.
\newblock


\bibitem[Goodfellow et~al\mbox{.}(2014)]%
        {goodfellow2014generative}
\bibfield{author}{\bibinfo{person}{Ian Goodfellow}, \bibinfo{person}{Jean
  Pouget-Abadie}, \bibinfo{person}{Mehdi Mirza}, \bibinfo{person}{Bing Xu},
  \bibinfo{person}{David Warde-Farley}, \bibinfo{person}{Sherjil Ozair},
  \bibinfo{person}{Aaron Courville}, {and} \bibinfo{person}{Yoshua Bengio}.}
  \bibinfo{year}{2014}\natexlab{}.
\newblock \showarticletitle{Generative adversarial nets}. In
  \bibinfo{booktitle}{\emph{NeurIPS}}.
\newblock


\bibitem[Gu et~al\mbox{.}(2022)]%
        {gu2021stylenerf}
\bibfield{author}{\bibinfo{person}{Jiatao Gu}, \bibinfo{person}{Lingjie Liu},
  \bibinfo{person}{Peng Wang}, {and} \bibinfo{person}{Christian Theobalt}.}
  \bibinfo{year}{2022}\natexlab{}.
\newblock \showarticletitle{StyleNeRF: A Style-based 3D-Aware Generator for
  High-resolution Image Synthesis}. In \bibinfo{booktitle}{\emph{ICLR}}.
\newblock


\bibitem[H{\"a}rk{\"o}nen et~al\mbox{.}(2020)]%
        {harkonen2020ganspace}
\bibfield{author}{\bibinfo{person}{Erik H{\"a}rk{\"o}nen},
  \bibinfo{person}{Aaron Hertzmann}, \bibinfo{person}{Jaakko Lehtinen}, {and}
  \bibinfo{person}{Sylvain Paris}.} \bibinfo{year}{2020}\natexlab{}.
\newblock \showarticletitle{GANSpace: Discovering Interpretable GAN Controls}.
\newblock \bibinfo{journal}{\emph{arXiv preprint arXiv:2004.02546}}
  (\bibinfo{year}{2020}).
\newblock


\bibitem[Harley et~al\mbox{.}(2022)]%
        {harley2022particle}
\bibfield{author}{\bibinfo{person}{Adam~W. Harley}, \bibinfo{person}{Zhaoyuan
  Fang}, {and} \bibinfo{person}{Katerina Fragkiadaki}.}
  \bibinfo{year}{2022}\natexlab{}.
\newblock \showarticletitle{Particle Video Revisited: {T}racking Through
  Occlusions Using Point Trajectories}. In \bibinfo{booktitle}{\emph{ECCV}}.
\newblock


\bibitem[Ho et~al\mbox{.}(2020)]%
        {ho2020denoising}
\bibfield{author}{\bibinfo{person}{Jonathan Ho}, \bibinfo{person}{Ajay Jain},
  {and} \bibinfo{person}{Pieter Abbeel}.} \bibinfo{year}{2020}\natexlab{}.
\newblock \showarticletitle{Denoising diffusion probabilistic models}. In
  \bibinfo{booktitle}{\emph{NeurIPS}}.
\newblock


\bibitem[Igarashi et~al\mbox{.}(2005)]%
        {igarashi2005rigid}
\bibfield{author}{\bibinfo{person}{Takeo Igarashi}, \bibinfo{person}{Tomer
  Moscovich}, {and} \bibinfo{person}{John~F Hughes}.}
  \bibinfo{year}{2005}\natexlab{}.
\newblock \showarticletitle{As-rigid-as-possible shape manipulation}.
\newblock \bibinfo{journal}{\emph{ACM transactions on Graphics (TOG)}}
  \bibinfo{volume}{24}, \bibinfo{number}{3} (\bibinfo{year}{2005}),
  \bibinfo{pages}{1134--1141}.
\newblock


\bibitem[Ilg et~al\mbox{.}(2017)]%
        {ilg2017flownet}
\bibfield{author}{\bibinfo{person}{Eddy Ilg}, \bibinfo{person}{Nikolaus Mayer},
  \bibinfo{person}{Tonmoy Saikia}, \bibinfo{person}{Margret Keuper},
  \bibinfo{person}{Alexey Dosovitskiy}, {and} \bibinfo{person}{Thomas Brox}.}
  \bibinfo{year}{2017}\natexlab{}.
\newblock \showarticletitle{Flownet 2.0: Evolution of optical flow estimation
  with deep networks}. In \bibinfo{booktitle}{\emph{CVPR}}.
\newblock


\bibitem[Isola et~al\mbox{.}(2017)]%
        {isola2017image}
\bibfield{author}{\bibinfo{person}{Phillip Isola}, \bibinfo{person}{Jun-Yan
  Zhu}, \bibinfo{person}{Tinghui Zhou}, {and} \bibinfo{person}{Alexei~A
  Efros}.} \bibinfo{year}{2017}\natexlab{}.
\newblock \showarticletitle{Image-to-image translation with conditional
  adversarial networks}. In \bibinfo{booktitle}{\emph{CVPR}}.
\newblock


\bibitem[Karras et~al\mbox{.}(2021)]%
        {Karras2021}
\bibfield{author}{\bibinfo{person}{Tero Karras}, \bibinfo{person}{Miika
  Aittala}, \bibinfo{person}{Samuli Laine}, \bibinfo{person}{Erik
  H\"ark\"onen}, \bibinfo{person}{Janne Hellsten}, \bibinfo{person}{Jaakko
  Lehtinen}, {and} \bibinfo{person}{Timo Aila}.}
  \bibinfo{year}{2021}\natexlab{}.
\newblock \showarticletitle{Alias-Free Generative Adversarial Networks}. In
  \bibinfo{booktitle}{\emph{NeurIPS}}.
\newblock


\bibitem[Karras et~al\mbox{.}(2019)]%
        {karras2019style}
\bibfield{author}{\bibinfo{person}{Tero Karras}, \bibinfo{person}{Samuli
  Laine}, {and} \bibinfo{person}{Timo Aila}.} \bibinfo{year}{2019}\natexlab{}.
\newblock \showarticletitle{A style-based generator architecture for generative
  adversarial networks}. In \bibinfo{booktitle}{\emph{CVPR}}.
  \bibinfo{pages}{4401--4410}.
\newblock


\bibitem[Karras et~al\mbox{.}(2020)]%
        {karras2020analyzing}
\bibfield{author}{\bibinfo{person}{Tero Karras}, \bibinfo{person}{Samuli
  Laine}, \bibinfo{person}{Miika Aittala}, \bibinfo{person}{Janne Hellsten},
  \bibinfo{person}{Jaakko Lehtinen}, {and} \bibinfo{person}{Timo Aila}.}
  \bibinfo{year}{2020}\natexlab{}.
\newblock \showarticletitle{Analyzing and improving the image quality of
  stylegan}. In \bibinfo{booktitle}{\emph{CVPR}}. \bibinfo{pages}{8110--8119}.
\newblock


\bibitem[King(2009)]%
        {dlib09}
\bibfield{author}{\bibinfo{person}{Davis~E. King}.}
  \bibinfo{year}{2009}\natexlab{}.
\newblock \showarticletitle{Dlib-ml: A Machine Learning Toolkit}.
\newblock \bibinfo{journal}{\emph{Journal of Machine Learning Research}}
  \bibinfo{volume}{10} (\bibinfo{year}{2009}), \bibinfo{pages}{1755--1758}.
\newblock


\bibitem[Kingma and Ba(2014)]%
        {kingma2014adam}
\bibfield{author}{\bibinfo{person}{Diederik~P Kingma} {and}
  \bibinfo{person}{Jimmy Ba}.} \bibinfo{year}{2014}\natexlab{}.
\newblock \showarticletitle{Adam: A method for stochastic optimization}.
\newblock \bibinfo{journal}{\emph{arXiv preprint arXiv:1412.6980}}
  (\bibinfo{year}{2014}).
\newblock


\bibitem[Leimk\"uhler and Drettakis(2021)]%
        {FreeStyleGAN2021}
\bibfield{author}{\bibinfo{person}{Thomas Leimk\"uhler} {and}
  \bibinfo{person}{George Drettakis}.} \bibinfo{year}{2021}\natexlab{}.
\newblock \showarticletitle{FreeStyleGAN: Free-view Editable Portrait Rendering
  with the Camera Manifold}.
\newblock  \bibinfo{volume}{40}, \bibinfo{number}{6} (\bibinfo{year}{2021}).
\newblock
\urldef\tempurl%
\url{https://doi.org/10.1145/3478513.3480538}
\showDOI{\tempurl}


\bibitem[Ling et~al\mbox{.}(2021)]%
        {ling2021editgan}
\bibfield{author}{\bibinfo{person}{Huan Ling}, \bibinfo{person}{Karsten Kreis},
  \bibinfo{person}{Daiqing Li}, \bibinfo{person}{Seung~Wook Kim},
  \bibinfo{person}{Antonio Torralba}, {and} \bibinfo{person}{Sanja Fidler}.}
  \bibinfo{year}{2021}\natexlab{}.
\newblock \showarticletitle{Editgan: High-precision semantic image editing}. In
  \bibinfo{booktitle}{\emph{NeurIPS}}.
\newblock


\bibitem[Lipman et~al\mbox{.}(2004)]%
        {lipman2004differential}
\bibfield{author}{\bibinfo{person}{Yaron Lipman}, \bibinfo{person}{Olga
  Sorkine}, \bibinfo{person}{Daniel Cohen-Or}, \bibinfo{person}{David Levin},
  \bibinfo{person}{Christian Rossi}, {and} \bibinfo{person}{Hans-Peter
  Seidel}.} \bibinfo{year}{2004}\natexlab{}.
\newblock \showarticletitle{Differential coordinates for interactive mesh
  editing}. In \bibinfo{booktitle}{\emph{Proceedings Shape Modeling
  Applications, 2004.}} IEEE, \bibinfo{pages}{181--190}.
\newblock


\bibitem[Lipman et~al\mbox{.}(2005)]%
        {lipman2005linear}
\bibfield{author}{\bibinfo{person}{Yaron Lipman}, \bibinfo{person}{Olga
  Sorkine}, \bibinfo{person}{David Levin}, {and} \bibinfo{person}{Daniel
  Cohen-Or}.} \bibinfo{year}{2005}\natexlab{}.
\newblock \showarticletitle{Linear rotation-invariant coordinates for meshes}.
\newblock \bibinfo{journal}{\emph{ACM Transactions on Graphics (ToG)}}
  \bibinfo{volume}{24}, \bibinfo{number}{3} (\bibinfo{year}{2005}),
  \bibinfo{pages}{479--487}.
\newblock


\bibitem[Mokady et~al\mbox{.}(2022)]%
        {mokady2022self}
\bibfield{author}{\bibinfo{person}{Ron Mokady}, \bibinfo{person}{Omer Tov},
  \bibinfo{person}{Michal Yarom}, \bibinfo{person}{Oran Lang},
  \bibinfo{person}{Inbar Mosseri}, \bibinfo{person}{Tali Dekel},
  \bibinfo{person}{Daniel Cohen-Or}, {and} \bibinfo{person}{Michal Irani}.}
  \bibinfo{year}{2022}\natexlab{}.
\newblock \showarticletitle{Self-distilled stylegan: Towards generation from
  internet photos}. In \bibinfo{booktitle}{\emph{ACM SIGGRAPH 2022 Conference
  Proceedings}}. \bibinfo{pages}{1--9}.
\newblock


\bibitem[Pan et~al\mbox{.}(2021)]%
        {pan2021shadegan}
\bibfield{author}{\bibinfo{person}{Xingang Pan}, \bibinfo{person}{Xudong Xu},
  \bibinfo{person}{Chen~Change Loy}, \bibinfo{person}{Christian Theobalt},
  {and} \bibinfo{person}{Bo Dai}.} \bibinfo{year}{2021}\natexlab{}.
\newblock \showarticletitle{A Shading-Guided Generative Implicit Model for
  Shape-Accurate 3D-Aware Image Synthesis}. In
  \bibinfo{booktitle}{\emph{NeurIPS}}.
\newblock


\bibitem[Park et~al\mbox{.}(2019)]%
        {park2019semantic}
\bibfield{author}{\bibinfo{person}{Taesung Park}, \bibinfo{person}{Ming-Yu
  Liu}, \bibinfo{person}{Ting-Chun Wang}, {and} \bibinfo{person}{Jun-Yan Zhu}.}
  \bibinfo{year}{2019}\natexlab{}.
\newblock \showarticletitle{Semantic image synthesis with spatially-adaptive
  normalization}. In \bibinfo{booktitle}{\emph{CVPR}}.
\newblock


\bibitem[Paszke et~al\mbox{.}(2017)]%
        {paszke2017automatic}
\bibfield{author}{\bibinfo{person}{Adam Paszke}, \bibinfo{person}{Sam Gross},
  \bibinfo{person}{Soumith Chintala}, \bibinfo{person}{Gregory Chanan},
  \bibinfo{person}{Edward Yang}, \bibinfo{person}{Zachary DeVito},
  \bibinfo{person}{Zeming Lin}, \bibinfo{person}{Alban Desmaison},
  \bibinfo{person}{Luca Antiga}, {and} \bibinfo{person}{Adam Lerer}.}
  \bibinfo{year}{2017}\natexlab{}.
\newblock \showarticletitle{Automatic differentiation in PyTorch}.
\newblock  (\bibinfo{year}{2017}).
\newblock


\bibitem[Patashnik et~al\mbox{.}(2021)]%
        {patashnik2021styleclip}
\bibfield{author}{\bibinfo{person}{Or Patashnik}, \bibinfo{person}{Zongze Wu},
  \bibinfo{person}{Eli Shechtman}, \bibinfo{person}{Daniel Cohen-Or}, {and}
  \bibinfo{person}{Dani Lischinski}.} \bibinfo{year}{2021}\natexlab{}.
\newblock \showarticletitle{Styleclip: Text-driven manipulation of stylegan
  imagery}. In \bibinfo{booktitle}{\emph{ICCV}}.
\newblock


\bibitem[Pinkney(2020)]%
        {Pinkney_Awesome_pretrained_StyleGAN2}
\bibfield{author}{\bibinfo{person}{Justin N.~M. Pinkney}.}
  \bibinfo{year}{2020}\natexlab{}.
\newblock \bibinfo{title}{{Awesome pretrained StyleGAN2}}.
\newblock
  \bibinfo{howpublished}{\url{https://github.com/justinpinkney/awesome-pretrained-stylegan2}}.
\newblock


\bibitem[Ramesh et~al\mbox{.}(2022)]%
        {ramesh2022hierarchical}
\bibfield{author}{\bibinfo{person}{Aditya Ramesh}, \bibinfo{person}{Prafulla
  Dhariwal}, \bibinfo{person}{Alex Nichol}, \bibinfo{person}{Casey Chu}, {and}
  \bibinfo{person}{Mark Chen}.} \bibinfo{year}{2022}\natexlab{}.
\newblock \showarticletitle{Hierarchical text-conditional image generation with
  clip latents}.
\newblock \bibinfo{journal}{\emph{arXiv preprint arXiv:2204.06125}}
  (\bibinfo{year}{2022}).
\newblock


\bibitem[Roich et~al\mbox{.}(2022)]%
        {roich2022pivotal}
\bibfield{author}{\bibinfo{person}{Daniel Roich}, \bibinfo{person}{Ron Mokady},
  \bibinfo{person}{Amit~H Bermano}, {and} \bibinfo{person}{Daniel Cohen-Or}.}
  \bibinfo{year}{2022}\natexlab{}.
\newblock \showarticletitle{Pivotal tuning for latent-based editing of real
  images}.
\newblock \bibinfo{journal}{\emph{ACM Transactions on Graphics (TOG)}}
  \bibinfo{volume}{42}, \bibinfo{number}{1} (\bibinfo{year}{2022}),
  \bibinfo{pages}{1--13}.
\newblock


\bibitem[Rombach et~al\mbox{.}(2021)]%
        {rombach2021highresolution}
\bibfield{author}{\bibinfo{person}{Robin Rombach}, \bibinfo{person}{Andreas
  Blattmann}, \bibinfo{person}{Dominik Lorenz}, \bibinfo{person}{Patrick
  Esser}, {and} \bibinfo{person}{Björn Ommer}.}
  \bibinfo{year}{2021}\natexlab{}.
\newblock \bibinfo{title}{High-Resolution Image Synthesis with Latent Diffusion
  Models}.
\newblock
\newblock
\showeprint[arxiv]{2112.10752}~[cs.CV]


\bibitem[Saharia et~al\mbox{.}(2022)]%
        {saharia2022photorealistic}
\bibfield{author}{\bibinfo{person}{Chitwan Saharia}, \bibinfo{person}{William
  Chan}, \bibinfo{person}{Saurabh Saxena}, \bibinfo{person}{Lala Li},
  \bibinfo{person}{Jay Whang}, \bibinfo{person}{Emily Denton},
  \bibinfo{person}{Seyed Kamyar~Seyed Ghasemipour},
  \bibinfo{person}{Burcu~Karagol Ayan}, \bibinfo{person}{S~Sara Mahdavi},
  \bibinfo{person}{Rapha~Gontijo Lopes}, {et~al\mbox{.}}}
  \bibinfo{year}{2022}\natexlab{}.
\newblock \showarticletitle{Photorealistic Text-to-Image Diffusion Models with
  Deep Language Understanding}.
\newblock \bibinfo{journal}{\emph{arXiv preprint arXiv:2205.11487}}
  (\bibinfo{year}{2022}).
\newblock


\bibitem[Schwarz et~al\mbox{.}(2020)]%
        {Schwarz2020NEURIPS}
\bibfield{author}{\bibinfo{person}{Katja Schwarz}, \bibinfo{person}{Yiyi Liao},
  \bibinfo{person}{Michael Niemeyer}, {and} \bibinfo{person}{Andreas Geiger}.}
  \bibinfo{year}{2020}\natexlab{}.
\newblock \showarticletitle{GRAF: Generative Radiance Fields for 3D-Aware Image
  Synthesis}. In \bibinfo{booktitle}{\emph{NeurIPS}}.
\newblock


\bibitem[Shen et~al\mbox{.}(2020)]%
        {shen2020interpreting}
\bibfield{author}{\bibinfo{person}{Yujun Shen}, \bibinfo{person}{Jinjin Gu},
  \bibinfo{person}{Xiaoou Tang}, {and} \bibinfo{person}{Bolei Zhou}.}
  \bibinfo{year}{2020}\natexlab{}.
\newblock \showarticletitle{Interpreting the latent space of gans for semantic
  face editing}. In \bibinfo{booktitle}{\emph{CVPR}}.
\newblock


\bibitem[Shen and Zhou(2020)]%
        {shen2020closed}
\bibfield{author}{\bibinfo{person}{Yujun Shen} {and} \bibinfo{person}{Bolei
  Zhou}.} \bibinfo{year}{2020}\natexlab{}.
\newblock \showarticletitle{Closed-Form Factorization of Latent Semantics in
  GANs}.
\newblock \bibinfo{journal}{\emph{arXiv preprint arXiv:2007.06600}}
  (\bibinfo{year}{2020}).
\newblock


\bibitem[Skorokhodov et~al\mbox{.}(2021)]%
        {ALIS}
\bibfield{author}{\bibinfo{person}{Ivan Skorokhodov}, \bibinfo{person}{Grigorii
  Sotnikov}, {and} \bibinfo{person}{Mohamed Elhoseiny}.}
  \bibinfo{year}{2021}\natexlab{}.
\newblock \showarticletitle{Aligning Latent and Image Spaces to Connect the
  Unconnectable}.
\newblock \bibinfo{journal}{\emph{arXiv preprint arXiv:2104.06954}}
  (\bibinfo{year}{2021}).
\newblock


\bibitem[Sohl-Dickstein et~al\mbox{.}(2015)]%
        {sohl2015deep}
\bibfield{author}{\bibinfo{person}{Jascha Sohl-Dickstein},
  \bibinfo{person}{Eric Weiss}, \bibinfo{person}{Niru Maheswaranathan}, {and}
  \bibinfo{person}{Surya Ganguli}.} \bibinfo{year}{2015}\natexlab{}.
\newblock \showarticletitle{Deep unsupervised learning using nonequilibrium
  thermodynamics}. In \bibinfo{booktitle}{\emph{International Conference on
  Machine Learning}}. PMLR, \bibinfo{pages}{2256--2265}.
\newblock


\bibitem[Song et~al\mbox{.}(2020)]%
        {song2020denoising}
\bibfield{author}{\bibinfo{person}{Jiaming Song}, \bibinfo{person}{Chenlin
  Meng}, {and} \bibinfo{person}{Stefano Ermon}.}
  \bibinfo{year}{2020}\natexlab{}.
\newblock \showarticletitle{Denoising Diffusion Implicit Models}. In
  \bibinfo{booktitle}{\emph{ICLR}}.
\newblock


\bibitem[Song et~al\mbox{.}(2021)]%
        {song2021scorebased}
\bibfield{author}{\bibinfo{person}{Yang Song}, \bibinfo{person}{Jascha
  Sohl-Dickstein}, \bibinfo{person}{Diederik~P Kingma},
  \bibinfo{person}{Abhishek Kumar}, \bibinfo{person}{Stefano Ermon}, {and}
  \bibinfo{person}{Ben Poole}.} \bibinfo{year}{2021}\natexlab{}.
\newblock \showarticletitle{Score-Based Generative Modeling through Stochastic
  Differential Equations}. In \bibinfo{booktitle}{\emph{International
  Conference on Learning Representations}}.
\newblock


\bibitem[Sorkine and Alexa(2007)]%
        {sorkine2007rigid}
\bibfield{author}{\bibinfo{person}{Olga Sorkine} {and} \bibinfo{person}{Marc
  Alexa}.} \bibinfo{year}{2007}\natexlab{}.
\newblock \showarticletitle{As-rigid-as-possible surface modeling}. In
  \bibinfo{booktitle}{\emph{Symposium on Geometry processing}},
  Vol.~\bibinfo{volume}{4}. Citeseer, \bibinfo{pages}{109--116}.
\newblock


\bibitem[Sorkine et~al\mbox{.}(2004)]%
        {sorkine2004laplacian}
\bibfield{author}{\bibinfo{person}{Olga Sorkine}, \bibinfo{person}{Daniel
  Cohen-Or}, \bibinfo{person}{Yaron Lipman}, \bibinfo{person}{Marc Alexa},
  \bibinfo{person}{Christian R{\"o}ssl}, {and} \bibinfo{person}{H-P Seidel}.}
  \bibinfo{year}{2004}\natexlab{}.
\newblock \showarticletitle{Laplacian surface editing}. In
  \bibinfo{booktitle}{\emph{Proceedings of the 2004 Eurographics/ACM SIGGRAPH
  symposium on Geometry processing}}. \bibinfo{pages}{175--184}.
\newblock


\bibitem[Sundaram et~al\mbox{.}(2010)]%
        {sundaram2010dense}
\bibfield{author}{\bibinfo{person}{Narayanan Sundaram}, \bibinfo{person}{Thomas
  Brox}, {and} \bibinfo{person}{Kurt Keutzer}.}
  \bibinfo{year}{2010}\natexlab{}.
\newblock \showarticletitle{Dense point trajectories by gpu-accelerated large
  displacement optical flow}. In \bibinfo{booktitle}{\emph{ECCV}}.
\newblock


\bibitem[Suzuki et~al\mbox{.}(2018)]%
        {suzuki2018spatially}
\bibfield{author}{\bibinfo{person}{Ryohei Suzuki}, \bibinfo{person}{Masanori
  Koyama}, \bibinfo{person}{Takeru Miyato}, \bibinfo{person}{Taizan Yonetsuji},
  {and} \bibinfo{person}{Huachun Zhu}.} \bibinfo{year}{2018}\natexlab{}.
\newblock \showarticletitle{Spatially controllable image synthesis with
  internal representation collaging}.
\newblock \bibinfo{journal}{\emph{arXiv preprint arXiv:1811.10153}}
  (\bibinfo{year}{2018}).
\newblock


\bibitem[Teed and Deng(2020)]%
        {teed2020raft}
\bibfield{author}{\bibinfo{person}{Zachary Teed} {and} \bibinfo{person}{Jia
  Deng}.} \bibinfo{year}{2020}\natexlab{}.
\newblock \showarticletitle{Raft: Recurrent all-pairs field transforms for
  optical flow}. In \bibinfo{booktitle}{\emph{ECCV}}.
\newblock


\bibitem[Tewari et~al\mbox{.}(2022)]%
        {tewari2022d3d}
\bibfield{author}{\bibinfo{person}{Ayush Tewari}, \bibinfo{person}{MalliKarjun
  B~R}, \bibinfo{person}{Xingang Pan}, \bibinfo{person}{Ohad Fried},
  \bibinfo{person}{Maneesh Agrawala}, {and} \bibinfo{person}{Christian
  Theobalt}.} \bibinfo{year}{2022}\natexlab{}.
\newblock \showarticletitle{Disentangled3D: Learning a 3D Generative Model with
  Disentangled Geometry and Appearance from Monocular Images}. In
  \bibinfo{booktitle}{\emph{CVPR}}.
\newblock


\bibitem[Tewari et~al\mbox{.}(2020)]%
        {tewari2020stylerig}
\bibfield{author}{\bibinfo{person}{Ayush Tewari}, \bibinfo{person}{Mohamed
  Elgharib}, \bibinfo{person}{Gaurav Bharaj}, \bibinfo{person}{Florian
  Bernard}, \bibinfo{person}{Hans-Peter Seidel}, \bibinfo{person}{Patrick
  P{\'e}rez}, \bibinfo{person}{Michael Zollhofer}, {and}
  \bibinfo{person}{Christian Theobalt}.} \bibinfo{year}{2020}\natexlab{}.
\newblock \showarticletitle{StyleRig: Rigging StyleGAN for 3D Control over
  Portrait Images}. In \bibinfo{booktitle}{\emph{CVPR}}.
\newblock


\bibitem[Tritrong et~al\mbox{.}(2021)]%
        {tritrong2021repurposing}
\bibfield{author}{\bibinfo{person}{Nontawat Tritrong},
  \bibinfo{person}{Pitchaporn Rewatbowornwong}, {and} \bibinfo{person}{Supasorn
  Suwajanakorn}.} \bibinfo{year}{2021}\natexlab{}.
\newblock \showarticletitle{Repurposing gans for one-shot semantic part
  segmentation}. In \bibinfo{booktitle}{\emph{Proceedings of the IEEE/CVF
  conference on computer vision and pattern recognition}}.
  \bibinfo{pages}{4475--4485}.
\newblock


\bibitem[Wang et~al\mbox{.}(2022b)]%
        {wang2022improving}
\bibfield{author}{\bibinfo{person}{Jianyuan Wang}, \bibinfo{person}{Ceyuan
  Yang}, \bibinfo{person}{Yinghao Xu}, \bibinfo{person}{Yujun Shen},
  \bibinfo{person}{Hongdong Li}, {and} \bibinfo{person}{Bolei Zhou}.}
  \bibinfo{year}{2022}\natexlab{b}.
\newblock \showarticletitle{Improving gan equilibrium by raising spatial
  awareness}. In \bibinfo{booktitle}{\emph{CVPR}}.
  \bibinfo{pages}{11285--11293}.
\newblock


\bibitem[Wang et~al\mbox{.}(2022a)]%
        {wang2022rewriting}
\bibfield{author}{\bibinfo{person}{Sheng-Yu Wang}, \bibinfo{person}{David Bau},
  {and} \bibinfo{person}{Jun-Yan Zhu}.} \bibinfo{year}{2022}\natexlab{a}.
\newblock \showarticletitle{Rewriting Geometric Rules of a GAN}.
\newblock \bibinfo{journal}{\emph{ACM Transactions on Graphics (TOG)}}
  (\bibinfo{year}{2022}).
\newblock


\bibitem[Xu et~al\mbox{.}(2022)]%
        {xu2022volumegan}
\bibfield{author}{\bibinfo{person}{Yinghao Xu}, \bibinfo{person}{Sida Peng},
  \bibinfo{person}{Ceyuan Yang}, \bibinfo{person}{Yujun Shen}, {and}
  \bibinfo{person}{Bolei Zhou}.} \bibinfo{year}{2022}\natexlab{}.
\newblock \showarticletitle{3D-aware Image Synthesis via Learning Structural
  and Textural Representations}. In \bibinfo{booktitle}{\emph{CVPR}}.
\newblock


\bibitem[Yu et~al\mbox{.}(2015)]%
        {yu2015lsun}
\bibfield{author}{\bibinfo{person}{Fisher Yu}, \bibinfo{person}{Ari Seff},
  \bibinfo{person}{Yinda Zhang}, \bibinfo{person}{Shuran Song},
  \bibinfo{person}{Thomas Funkhouser}, {and} \bibinfo{person}{Jianxiong Xiao}.}
  \bibinfo{year}{2015}\natexlab{}.
\newblock \showarticletitle{Lsun: Construction of a large-scale image dataset
  using deep learning with humans in the loop}.
\newblock \bibinfo{journal}{\emph{arXiv preprint arXiv:1506.03365}}
  (\bibinfo{year}{2015}).
\newblock


\bibitem[Zhang et~al\mbox{.}(2018)]%
        {zhang2018perceptual}
\bibfield{author}{\bibinfo{person}{Richard Zhang}, \bibinfo{person}{Phillip
  Isola}, \bibinfo{person}{Alexei~A Efros}, \bibinfo{person}{Eli Shechtman},
  {and} \bibinfo{person}{Oliver Wang}.} \bibinfo{year}{2018}\natexlab{}.
\newblock \showarticletitle{The Unreasonable Effectiveness of Deep Features as
  a Perceptual Metric}. In \bibinfo{booktitle}{\emph{CVPR}}.
\newblock


\bibitem[Zhang et~al\mbox{.}(2021)]%
        {zhang21}
\bibfield{author}{\bibinfo{person}{Yuxuan Zhang}, \bibinfo{person}{Huan Ling},
  \bibinfo{person}{Jun Gao}, \bibinfo{person}{Kangxue Yin},
  \bibinfo{person}{Jean-Francois Lafleche}, \bibinfo{person}{Adela Barriuso},
  \bibinfo{person}{Antonio Torralba}, {and} \bibinfo{person}{Sanja Fidler}.}
  \bibinfo{year}{2021}\natexlab{}.
\newblock \showarticletitle{DatasetGAN: Efficient Labeled Data Factory with
  Minimal Human Effort}. In \bibinfo{booktitle}{\emph{CVPR}}.
\newblock


\bibitem[Zhu et~al\mbox{.}(2023)]%
        {zhu2023linkgan}
\bibfield{author}{\bibinfo{person}{Jiapeng Zhu}, \bibinfo{person}{Ceyuan Yang},
  \bibinfo{person}{Yujun Shen}, \bibinfo{person}{Zifan Shi},
  \bibinfo{person}{Deli Zhao}, {and} \bibinfo{person}{Qifeng Chen}.}
  \bibinfo{year}{2023}\natexlab{}.
\newblock \showarticletitle{LinkGAN: Linking GAN Latents to Pixels for
  Controllable Image Synthesis}.
\newblock \bibinfo{journal}{\emph{arXiv preprint arXiv:2301.04604}}
  (\bibinfo{year}{2023}).
\newblock


\bibitem[Zhu et~al\mbox{.}(2016)]%
        {zhu2016generative}
\bibfield{author}{\bibinfo{person}{Jun-Yan Zhu}, \bibinfo{person}{Philipp
  Kr{\"a}henb{\"u}hl}, \bibinfo{person}{Eli Shechtman}, {and}
  \bibinfo{person}{Alexei~A Efros}.} \bibinfo{year}{2016}\natexlab{}.
\newblock \showarticletitle{Generative visual manipulation on the natural image
  manifold}. In \bibinfo{booktitle}{\emph{ECCV}}.
\newblock


\end{thebibliography}

\begin{figure*}[h]
	\centering
	\includegraphics[width=14cm]{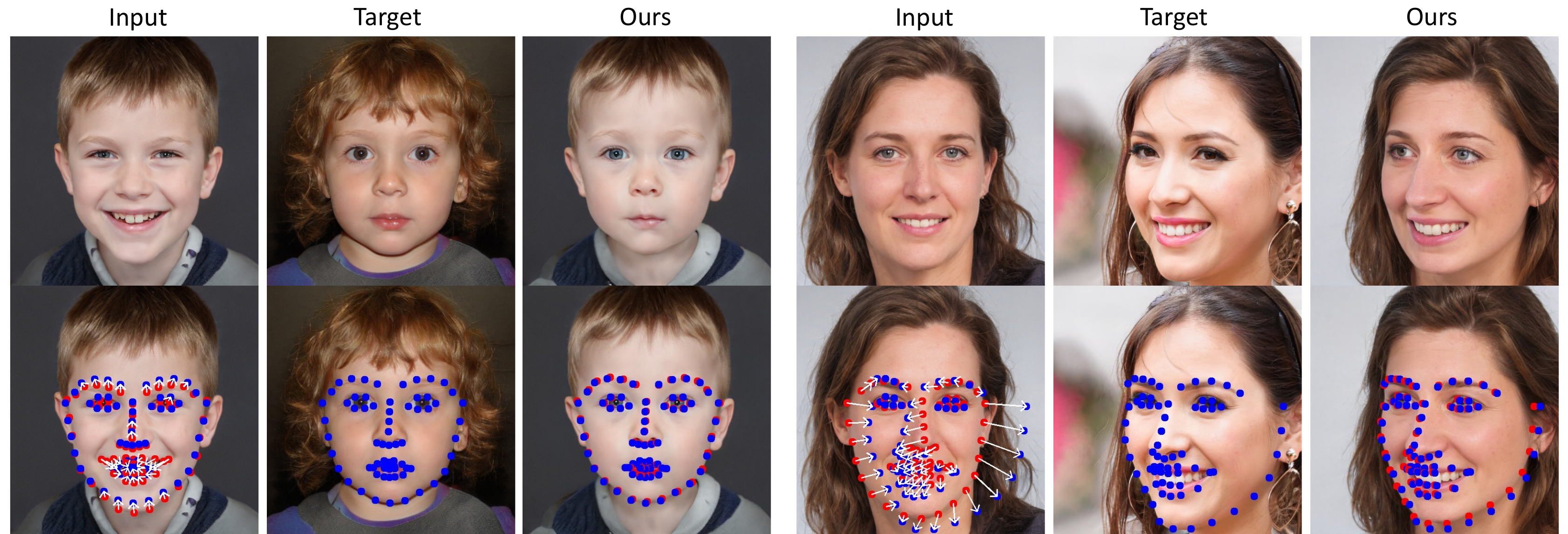}
    \vspace{-0.35cm}
	\caption{Face landmark manipulation. Our method works well even for such dense keypoint cases.}
	\label{fig:dense}
\end{figure*}

\begin{figure*}[h]
	\centering
	\includegraphics[width=16cm]{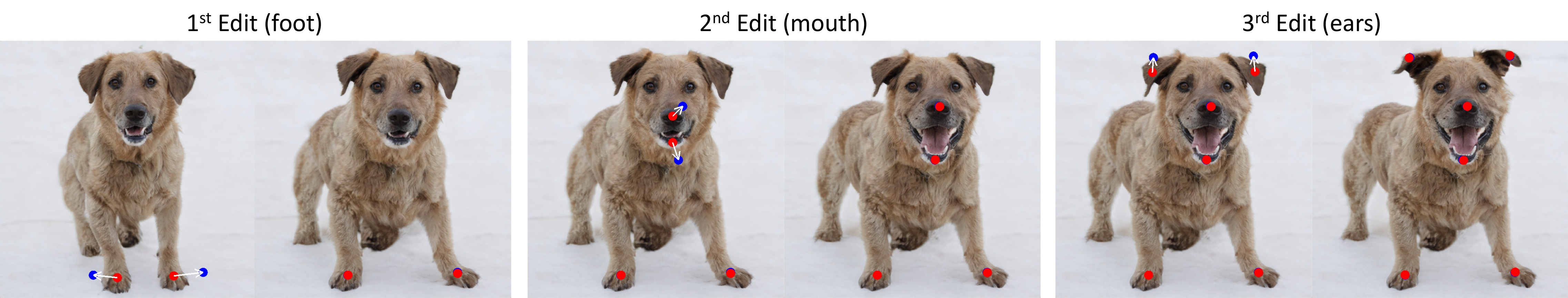}
    \vspace{-0.2cm}
	\caption{Continuous image manipulation. Users can continue the manipulation based on previous manipulation results.  }
	\label{fig:multiedit}
\end{figure*}

\begin{figure*}[h]
	\centering
	\includegraphics[width=15cm]{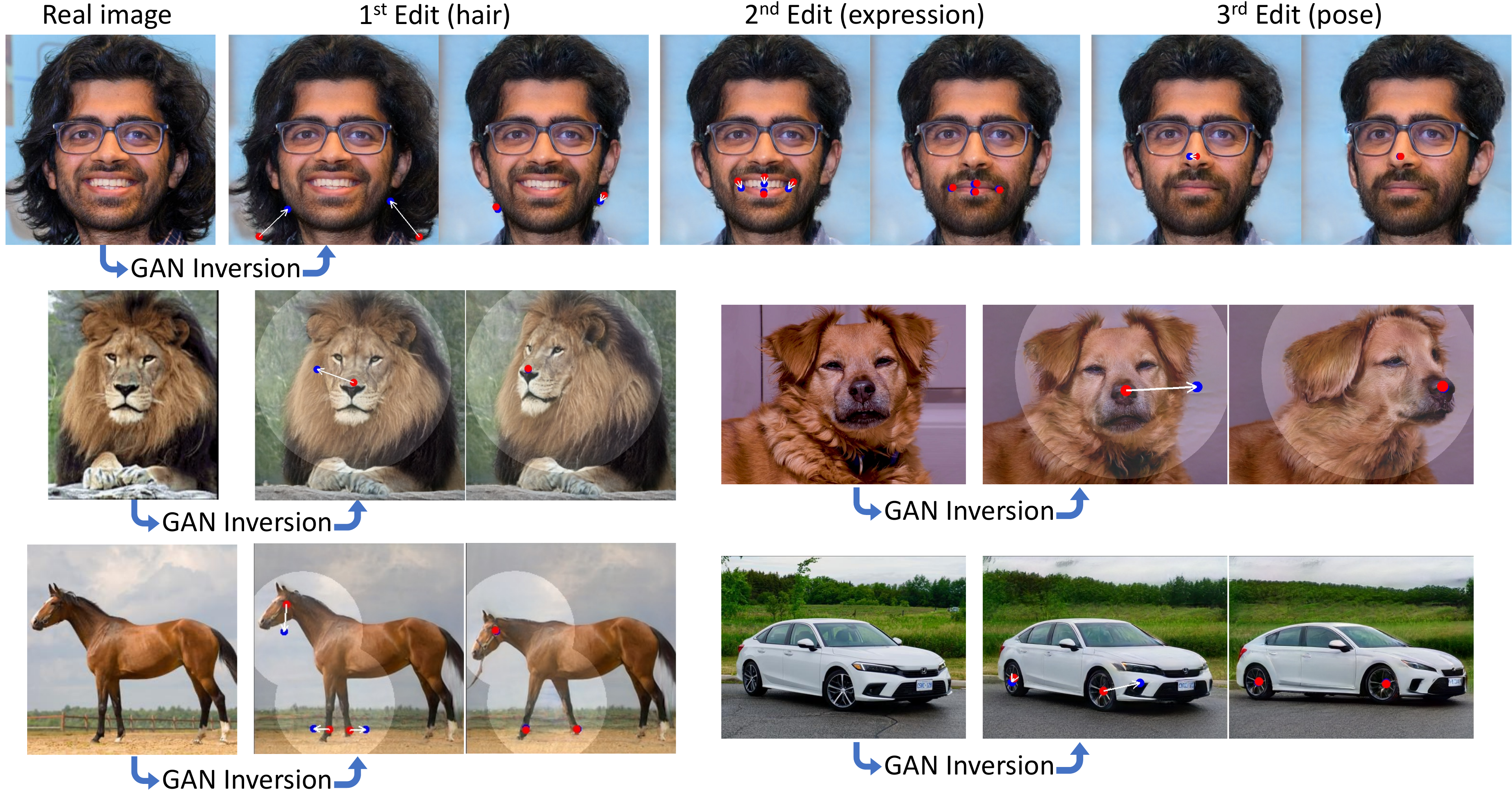}
    \vspace{-0.5cm}
	\caption{Real image manipulation. }
	\label{fig:real}
\end{figure*}

\begin{figure*}[h]
	\centering
	\includegraphics[width=\textwidth]{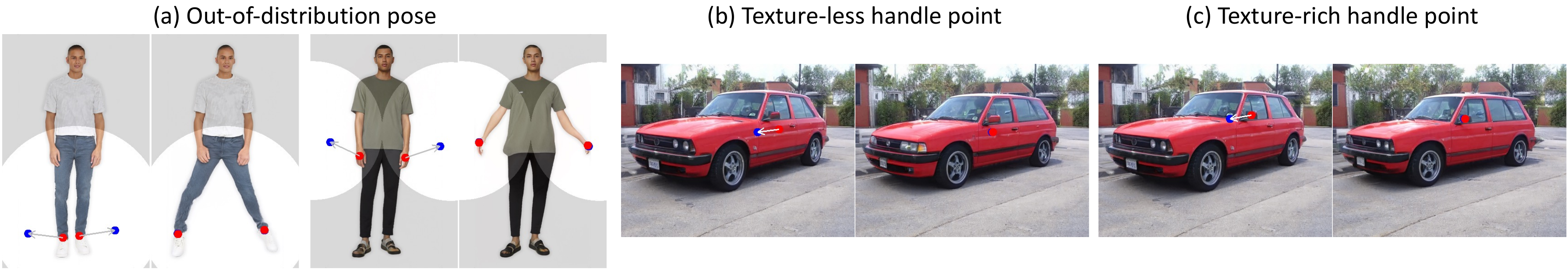}
    \vspace{-0.7cm}
	\caption{Limitations. (a) the StyleGAN-human~\cite{fu2022styleganhuman} is trained on a fashion dataset where most arms and legs are downward. Editing toward out-of-distribution poses can cause distortion artifacts as shown in the legs and hands. (b)\&(c) The handle point (\textcolor{red}{red}) in texture-less regions may suffer from more drift during tracking, as can be observed from its relative position to the rearview mirror. }
	\label{fig:limitation}
\end{figure*}

\begin{figure*}[h]
	\centering
 	\begin{minipage}{0.54\linewidth}
		\centering
	\includegraphics[width=9.7cm]{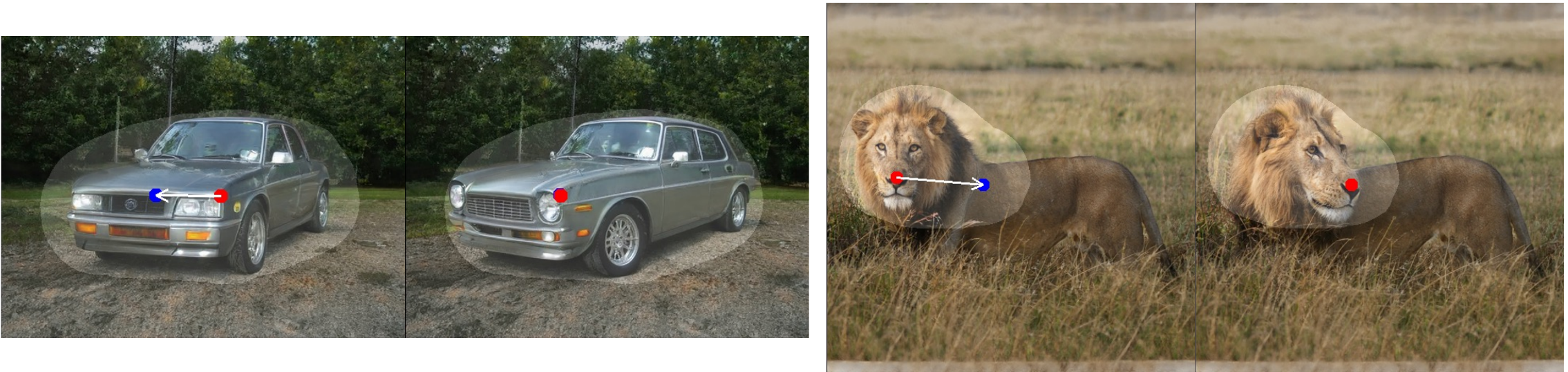}
\vspace{-0.6cm}
\caption{
Effects of the mask. By masking the foreground object, we can fix the background. The details of the trees and grasses are kept nearly unchanged. Better background preservation could potentially be achieved via feature blending~\cite{suzuki2018spatially}.
}
\label{fig:background}
	\end{minipage}
        \hfill
	\begin{minipage}{0.44\linewidth}
	\centering
\includegraphics[width=7.7cm]{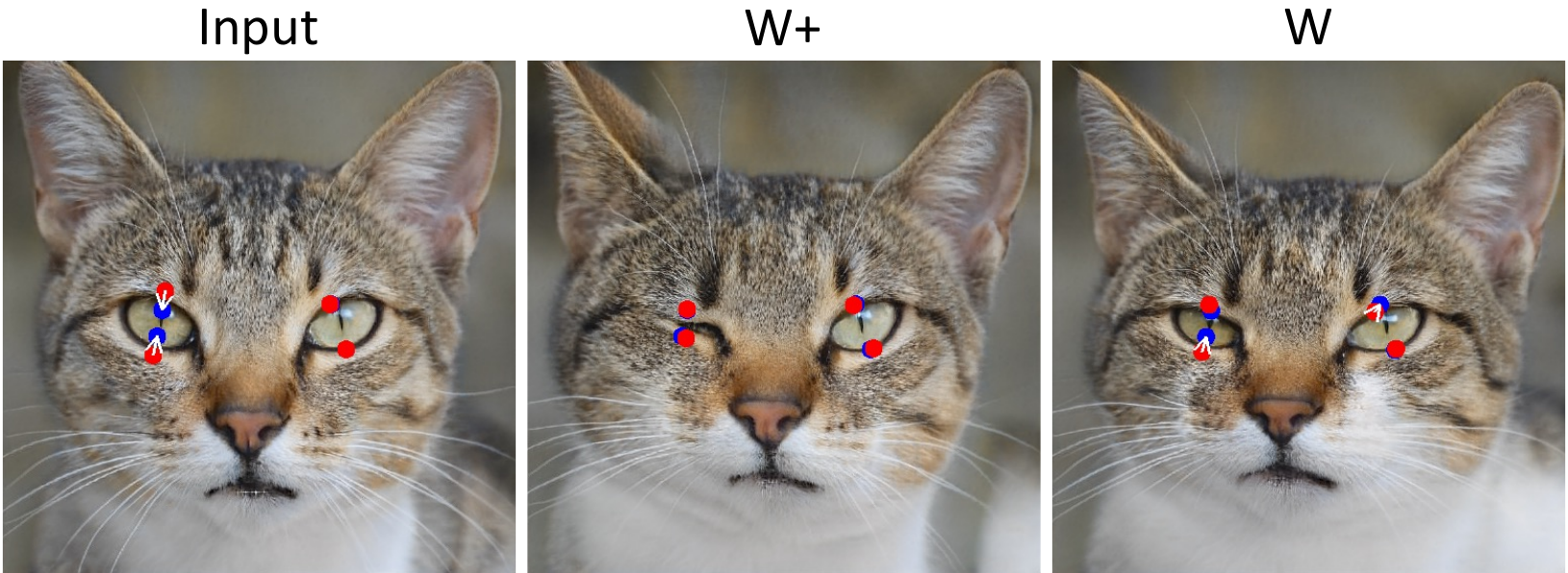}
\vspace{-0.35cm}
\caption{Effects of $\mathcal{W}$/$\mathcal{W}^+$ space. Optimizing the latent code in $\mathcal{W+}$ space is easier to achieve out-of-distribution manipulations such as closing only one eye of the cat. In contrast, $\mathcal{W}$ space struggles to achieve this as it tends to keep the image within the distribution of training data. }
\vspace{-0.35cm}
\label{fig:W}
	\end{minipage}
\end{figure*}



\end{document}